%% file: main.tex
\documentclass[conference,compsoc]{IEEEtran}
 \IEEEoverridecommandlockouts

\input{preamble}
\begin{document}

\title{SoK: A Systematic Evaluation of Backdoor Trigger Characteristics in Image Classification}

\author{
    \IEEEauthorblockN{Gorka Abad~*}
    \IEEEauthorblockA{\textit{Radboud University}\\
    \textit{Ikerlan research centre}\\
    \textit{abad.gorka@ru.nl}
    \thanks{* means equal contribution.}
    }\\
    \IEEEauthorblockN{Behrad Tajalli~*}
    \IEEEauthorblockA{\textit{Radboud University}\\
    \textit{hamidreza.tajalli@ru.nl}
    }
    \and
    \IEEEauthorblockN{Jing Xu~*}
    \IEEEauthorblockA{\textit{Delft University of Technology}\\
    \textit{j.xu-8@tudelft.nl}
    } \\
    \IEEEauthorblockN{Stjepan Picek}
    \IEEEauthorblockA{\textit{Radboud University}\\
    \textit{Delft University of Technology}\\
    \textit{stjepan.picek@ru.nl}
    }
    \and
    \IEEEauthorblockN{Stefanos Koffas~*}
    \IEEEauthorblockA{\textit{Delft University of Technology}\\
    \textit{s.koffas@tudelft.nl}
    }\\
    \IEEEauthorblockN{Mauro Conti}
    \IEEEauthorblockA{\textit{University of Padua}\\
    \textit{mauro.conti@unipd.it}
    }
}

\maketitle

\begin{abstract}

Deep learning achieves outstanding results in many machine learning tasks. Nevertheless, it is vulnerable to backdoor attacks that modify the training set to embed a secret functionality in the trained model. The modified training samples have a secret property, i. e., a trigger. At inference time, the secret functionality is activated when the input contains the trigger, while the model functions correctly in other cases. While there are many known backdoor attacks (and defenses), deploying a stealthy attack is still far from trivial. Successfully creating backdoor triggers depends on numerous parameters. Unfortunately, research has not yet determined which parameters contribute most to the attack performance.

This paper systematically analyzes the most relevant parameters for the backdoor attacks, i.e., trigger size, position, color, and poisoning rate. Using transfer learning, which is very common in computer vision, we evaluate the attack on state-of-the-art models (ResNet, VGG, AlexNet, and GoogLeNet) and datasets (MNIST, CIFAR10, and TinyImageNet). Our attacks cover the majority of backdoor settings in research, providing concrete directions for future works. Our code is publicly available\footnote{Code will be shared after paper acceptance.} to facilitate the reproducibility of our results.

\end{abstract}


\input{sections/1_introduction}

\input{sections/2_background}
\input{sections/2_motivation}
\input{sections/3_threat_model}
\input{sections/4_experiments}
\input{sections/_defenses}
\input{sections/5_related_work.tex}
\input{sections/6_conclusions}

\bibliography{bibliography}
\bibliographystyle{IEEEtran}

\input{sections/7_appendix}
\end{document}

%% file: preamble.tex
\usepackage{amsmath,amsthm,amsfonts}
\usepackage{algorithmic}
\usepackage{graphicx}
\usepackage{textcomp}
\usepackage{xcolor}
\usepackage[utf8]{inputenc}
\usepackage[normalem]{ulem}
\usepackage{xcolor}
\usepackage{color,colortbl}
\usepackage{lineno}
\usepackage{subcaption}
\usepackage{paralist}
\usepackage[hidelinks]{hyperref}
\usepackage[nameinlink]{cleveref}
\usepackage{booktabs}
\usepackage{multirow}
\usepackage{diagbox}
\usepackage{MnSymbol}
\usepackage{tikz}
\usepackage{tablefootnote}
\usepackage{xurl}

\newcommand\blackbox{\rule[0.2pt]{7pt}{7pt}}

\DeclareRobustCommand\whitesquare{
    \begin{tikzpicture}
        \draw[draw=black,fill=white] (0,0) rectangle (7pt,7pt);
    \end{tikzpicture}
}

\DeclareFontEncoding{LS2}{}{\noaccents@}
\DeclareFontSubstitution{LS2}{stix}{m}{n}
\DeclareSymbolFont{arrows3}{LS2}{stixtt}{m}{n}
\DeclareMathSymbol{\squarelrblack}{\mathord}{arrows3}{"89}

\theoremstyle{remark}
\usepackage{mdframed}
\newmdtheoremenv{remark}{Finding}

%% file: sections/1_introduction.tex
\section{Introduction}
\label{sec:introduction}

Deep neural networks (DNNs) have gained significant popularity over the past decade due to their performance in various application domains, including computer vision~\cite{deep-residual-learning-for-image-recognition}, speech recognition~\cite{speech-recongnition-with-deep-rnns}, and neural translation~\cite{google-translate}. 
One of the key benefits of DNNs is their ability to automatically learn and extract features from raw data, which reduces the need for manual feature engineering and makes them particularly well-suited for tasks where the data is complex or unstructured, such as image and audio processing~\cite{lecun2015deep}. Additionally, DNNs can efficiently process large amounts of data, achieving state-of-the-art performance on various tasks.
However, DNNs also have some limitations. For example, they require a large amount of labeled training data to perform well~\cite{imagenet-a-large-scale-hierarchical-image-database}, and they can be prone to overfitting if not adequately regularized~\cite{avoid-overfitting-a-survey-on-regularization}.
They also require significant computational resources and can be challenging to interpret due to their complex decision-making processes~\cite{selvaraju2017grad}.

The high computational requirements for training DNNs have led to emerging trends such as outsourced training and machine learning as a service~\cite{badnets-evaluating-backdoor-attacks-on-dnns}. These trends have introduced new threats for deployed models when they are provided as black boxes by third parties. In addition, malicious data samples can be easily embedded in widely-used crowdsourced datasets~\cite{large-dataset-pyrrhic-win}.
One approach to address the high computational requirements for training is \emph{transfer learning}, which involves using pre-trained models as a starting point for training on a new task~\cite{torrey2010transfer}. This can significantly reduce the amount of labeled training data and computational resources needed, as the pre-trained model has already learned many general features useful for diverse tasks. This has made transfer learning an essential tool in developing DNNs, particularly when labeled training data is limited or expensive.
In addition to the transfer learning, there have also been efforts to improve the interpretability of DNNs~\cite{selvaraju2017grad, ribeiro2016should}. This is important for various reasons, including the need to understand how a model makes decisions, the ability to identify and correct errors, and the development of trust in the model's outputs. One approach to improve interpretability uses visualization techniques, which can provide insight into the internal workings of a DNN and help identify patterns in the data the model is learning~\cite{selvaraju2017grad}.

Overall, DNNs have shown impressive performance on a wide range of tasks. 
Still, much work is needed to address their limitations and improve their interpretability and robustness.
One particularly concerning threat is the backdoor attack, which results in targeted misclassifications when a specific trigger is present in the input. Backdoor attacks can be mounted through data poisoning~\cite{badnets-evaluating-backdoor-attacks-on-dnns}, code poisoning~\cite{blind-backdoors-in-deep-learning-models}, or model poisoning~\cite{handcrafted-backdoors-in-dnns}. There has been a significant amount of research on backdoor attacks and their defenses in the literature~\cite{backdoor-attacks-and-countermeasures-on-deep-learning-a-comprehensive-review,backdoor-learning-a-survey}. Still, these works are empirical, based on prior assumptions, and not covering a wide range of the backdoors' parameter space.

Our paper focuses on the intersection between computer vision for image classification and data poisoning, the most common setup for mounting backdoor attacks. In particular, we systematically evaluate the impact of various parameters on the performance of various backdoor attacks. 
Our work extends previous research in this area~\cite{systematic-evaluation-of-backdoor-data-poisoning-attacks-on-image-classifiers} by using larger datasets with higher-dimensional images (we upsampled the images from MNIST to $64\times64$, from CIFAR10 to $128\times128$, and TinyImageNet to $224\times224$) and more classes. We also consider different state-of-the-art backdoor attacks and defenses. See~\autoref{sec:related_work},~\autoref{tab:comparison-related-work}, and~\autoref{tab:comparison-effect} for a detailed explanation of the differences with previous works. We find that the trigger size is more influential than the poisoning rate and that the performance of backdoor attacks is affected by factors such as the model architecture and the characteristics of the trigger. Finally, we demonstrate that AlexNet is more robust against data-poisoning backdoor attacks, and we conduct experiments to explain this finding.

Our main contributions are summarized as follows: 
\begin{compactitem} 
    \item We extend the work described in~\cite{systematic-evaluation-of-backdoor-data-poisoning-attacks-on-image-classifiers} by exploring a more comprehensive range of factors affecting backdoor performance. We also experiment with different state-of-the-art backdoor attacks. This allows us to provide findings that generalize well to the tested datasets and models representing state-of-the-art.
    \item Based on the extensive experimentation, we extract 1) \\dataset/model-specific and 2) general findings, which provide valuable insights for understanding the backdoor effect while easing the design of new attacks and defenses.
    \item We demonstrate that the performance of backdoor attacks is affected by various factors, including the model architecture and the characteristics of the trigger. 
    \item We show AlexNet is more robust against data poisoning backdoor attacks and conduct experiments to explain it.
    \item We additionally experiment with state-of-the-art defenses, evaluating the viability of the attacks in real-life scenarios.
\end{compactitem}





%% file: sections/2_background.tex
\section{Background}
\label{sec:background}

\subsection{Deep Neural Networks (DNNs)}

Deep learning algorithms are parameterized functions $\mathbb{F}_\theta$ that map an input $\textbf{x} \in \mathbb{R}^N$ to some output $y \in \mathbb{R}^M$. $\theta$ represents the parameters of the function, which are optimized via an iterative process called training. In the image domain, $\textbf{x}$ is an image, represented as a vector of pixel values, while $y$ is the vector of probabilities of the image being of a class $c \in k$ from a group of classes $k$. 
For training, a dataset is needed, i.e., a set of labeled samples $\mathcal{D} = \{\textbf{x}, y\}^n$ of size $n$. During training, the algorithm tries to find the optimal parameters $\theta'$ by minimizing the ``distance'' from the predicted labels to the ground truth ones. The distance calculation is done leveraging a loss function $\mathcal{L}$, which penalizes the algorithm depending on how far the prediction is from the actual label:
\begin{equation*}
      \theta' = \underset{\theta}{\mathrm{argmin}} \sum_{i=1}^n \mathcal{L} (\mathbb{F}_\theta (\{\textbf{x}_i, y_i \})).  
\end{equation*}

A convolutional neural network (CNN) performs convolutions in the input extracting relevant features linked to fully connected layers. The key intuition is to reduce the input space without losing information, which is easier to process in the consequent layers. This is achieved by kernels that move horizontally and vertically in the input in steps of a predefined value (stride). By doing so, the kernel extracts high-level representations as corners, shapes, or edges. Additionally, CNNs are accompanied by pooling layers that further reduce the computational complexity, extract the most relevant features, and reduce any noise captured by the kernels.



\subsection{Transfer Learning}

Transfer learning is adapting a pre-trained DNN to a related task without retraining the entire model from scratch~\cite{torrey2010transfer}. It involves adjusting the model's parameters, typically the fully connected layers, while freezing the convolutional layers in the case of CNNs. This approach offers cost-effective benefits when labeled training data is limited or expensive~\cite{badnets-evaluating-backdoor-attacks-on-dnns}. Transfer learning has wide applications in fields such as computer vision~\cite{sharif2014cnn}, natural language processing~\cite{howard2018universal}, and speech recognition~\cite{pan2010survey}, allowing DNNs to leverage previous knowledge for efficient learning. Factors such as task similarity, available labeled data, and feature reuse influence transfer learning's effectiveness~\cite{9134370}. It is a powerful tool for adapting DNNs to new tasks, especially in scenarios with limited labeled data.

\subsection{Backdoor Attacks in DNNs}

Backdoor attacks compromise DNNs during training and embed a secret functionality in the deployed model. This secret can be embedded through data poisoning~\cite{badnets-evaluating-backdoor-attacks-on-dnns,targeted-backdoor-attacks-on-deep-learning-systems-using-data-poisoning}, code poisoning~\cite{blind-backdoors-in-deep-learning-models}, or direct modification of the model's weights~\cite{handcrafted-backdoors-in-dnns}. In this work, we follow data poisoning by injecting poisoned samples into the training set. A poisoned sample contains a trigger, and its label is usually altered to the target label, which is the output of the model when the backdoor is activated.
In the image domain, the trigger is usually (but not limited to) a pixel pattern of a given color, e.g., white or black, placed anywhere over the image, creating a set of poisoned samples $\hat{\textbf{x}} \in D_{poison}$. The percentage of poisoned samples in the training set is controlled by $\epsilon = \frac{m}{n}$ where $m$ is the number of poisoned samples, $n$ is the number of the original training set and $ m\ll n$. A small $\epsilon$ makes the backdoor harder to embed but keeps it stealthier, as the small number of poisoned samples will not affect the original task much. A large $\epsilon$ leads to a stronger backdoor, but it could affect the original task substantially, making it somewhat unrealistic. During training with poisoned samples, the backdoor effect is included following:

\begin{equation*}
      \theta' = \underset{\theta}{\mathrm{argmin}} \sum_{i=1}^n \mathcal{L} (\mathbb{F}_\theta (\{\textbf{x}_i, y_i \})) + \sum_{j=1}^m \mathcal{L} (\mathbb{F}_\theta (\{\hat{\textbf{x}}_j, \hat{y}_j \})).  
\end{equation*}

After training, the backdoor is embedded in the DNN. The DNN functions normally on clean inputs, but the backdoor is activated in the presence of the trigger.

As stated, the backdoor trigger is usually a pixel pattern placed on the clean image. We refer to this as the \emph{BadNets attack}. However, more advanced attacks have also been developed. For instance, we additionally focus on SSBA~\cite{li2021invisible} and WaNet~\cite{wanet-imperceptible-warping-based-backdoor-attack}---see Section~\ref{sec:motivation}. SSBA leverages an encoder-decoder model as in~\cite{2019stegastamp} to add invisible perturbations in the images. The encoder-decoder takes an image and an attacker-defined string and encodes it, resulting in a slightly perturbed image. Note that the perturbation is sample-specific. Following the standard backdoor training, the malicious behavior gets injected into the model.
WaNet generates a trigger for images through a two-stage process. First, it creates a warping field using a normalized, upsampled, and clipped random tensor. Second, it trains the network with three modes: Attack (poisoning 10\% of samples with warping), Noise (adding warping and Gaussian noise to 20\% of samples, keeping labels unchanged), and Clean (training the remaining dataset without modification).

\subsection{On Backdoor Interpretability}

Interpretability techniques are used to explain the behavior of ML models. The interpretability of DNNs refers to the ability to understand the network's decision, which can be obtained by different methods, such as feature visualization. Typically, there is a trade-off between accuracy, simplicity, and explainability. For instance, shallow models such as linear regression or decision trees are highly interpretable~\cite{nguyen2020input, salem2022dynamic}.
By using DL models, we sacrifice the interpretability to achieve better performance, which often increases the complexity of the model by adding more layers. 

Recently, class activation mapping (CAM) was developed for CNNs, which identifies the regions of an image that are more linked to the model's prediction~\cite{zhou2016learning}. CAM modifies the architecture of the target model by changing the convolutional layers for fully connected layers, which are much more interpretable but incur a severe degradation in accuracy. Consequent work from Selvaraju et al. introduced a generalization of CAM called gradient-weighted CAM (Grad-CAM)~\cite{selvaraju2017grad}. Instead of modifying the model's architecture, it uses the gradient of a given class to produce a localization of the important regions of the image. Precisely, Grad-CAM computes the target class's gradients concerning the feature map activation of a convolutional layer.

Post-hoc interpretation methods have also been developed to explain individual predictions made by a DNN. These methods are applied after the model has been trained and do not require changes to the model's architecture or training process. One example is LIME (Local Interpretable Model-Agnostic Explanations)~\cite{ribeiro2016should}. LIME generates explanations by fitting a simple, interpretable model to the predictions made by a DNN in the vicinity of a particular input, allowing the model's behavior to be understood locally. 

Other post-hoc interpretation methods are SHAP~\cite{lundberg2017unified} and DeepLIFT~\cite{shrikumar2017learning}. SHAP uses Shapley values, a concept from game theory, to attribute the prediction made by a DNN to the individual features of the input. DeepLIFT computes the contribution of each feature to the final prediction by comparing the model's output with a reference score, which the user can choose.

Overall, various approaches are available for explaining ML models' behavior, including visualization techniques and post-hoc interpretation methods. The choice of method will depend on the specific requirements and constraints of the task. In this work, we will use Grad-CAM to understand the decisions of the poisoned models and compare their behavior with their clean counterparts, which is a suitable method for understanding the backdoor behavior~\cite{salem2022dynamic}.

%% file: sections/2_motivation.tex
\subsection{Motivation}
\label{sec:motivation}

In recent years, DL has become an extremely popular and rapidly evolving domain as a form to solve various real-world problems. Due to the need for adaptation to other tasks, DNNs have become more complex, often viewed as ``black boxes''. Indeed, Gilpin et al.~\cite{gilpin2018explaining} established a direct relationship between the models' complexity and their (lack of) explainability. Furthermore, efforts to create more complex and explainable models have been ongoing within the research community~\cite{aytekin2022neural}. At the same time, DNNs have also gained rising attention in the security community due to their vast applicability and impact of DNNs. The ability to understand and explain the inner workings of these models becomes particularly important in the context of DL attacks, as a lack of explainability can hinder our understanding of the root cause of security problems.

One type of DL attack that has garnered significant attention is the backdoor attack. Indeed, it has been recently subject to a deep investigation in a wide range of domains~\cite{backdoor-learning-a-survey}. In image recognition, the proposed attacks are heterogeneous in the trigger generation, backdoor injection, or threat model. Thus, comparing these attacks is far from trivial, even impossible in some cases. For instance, the models, datasets, experimental setups, and attack parameters are only a few to consider for comparing the performance of different attacks. Furthermore, even if the attacks are comparable, understanding the influence of the attack's parameters on the backdoor performance could still be difficult.

In this paper, we aim to address these issues by systematically investigating the impact of common parameters on the effectiveness of backdoor attacks in clean and backdoor performance. We analyze the core group of backdoor attacks in image classification, where the rest of the attacks build upon. For that, we investigated the backdoor attacks literature in the image domain, from which we selected the most representative attacks. Namely, BadNets~\cite{badnets-evaluating-backdoor-attacks-on-dnns}, SSBA~\cite{li2021invisible}, and WaNet~\cite{wanet-imperceptible-warping-based-backdoor-attack}. Then, we further investigated papers in the literature that fall into one of the categories above. The papers that fulfilled our criteria are shown in~\autoref{tab:comparison}. By analyzing those, we found an inconsistency in the parameter selection and the understanding of these parameters' effect on the backdoor performance. Thus, we propose systematically analyzing the attack proposals based on the same parameters and investigating the influence of these parameters on the main and backdoor task's performance. This allows us to efficiently and systematically compare a new attack---allowing fair and traceable comparisons. Our final goal is to provide a comprehensive and systematic analysis of the impact of parameters on backdoor attack performance. By doing so, we hope to contribute to a better understanding of these types of attacks and provide a valuable investigation for comparing and evaluating future research in this area.

For our investigation, we provide a realistic attack configuration, and we have designed our experimental setup to be as simple as possible while still being extendable to future backdoor attacks. To this end, we have surveyed the state-of-the-art to identify a common set of experimental settings that can be used to compare and evaluate future attacks. In line with previous research~\cite{backdoor-learning-a-survey}, we have chosen to focus on image recognition, using the MNIST, CIFAR10, and TinyImageNet datasets and models AlexNet, ResNet, VGG, and GoogLeNet. These datasets and models are representative samples of the ones used in the state-of-the-art.

Additionally, it is important to note that the choice of parameters can significantly impact the performance of a backdoor attack. For example, the trigger size, poisoning rate, and type of trigger can affect the attack's success rate. Similarly, the choice of dataset and model architecture can significantly affect the attack's effectiveness. By considering various values of parameters in our experiments, we explored a range of possibilities and identified differences in attack performance that may be related to each parameter. This is important in understanding the underlying mechanisms of backdoor attacks and developing more effective countermeasures. 



\begin{table}[htb]
\centering
\caption{Comparison of the attack setting for different state-of-the-art backdoor attacks that use patch triggers.}
\label{tab:comparison}
\resizebox{\columnwidth}{!}{%
\begin{tabular}{@{}ccccc@{}}
\toprule
Paper &
  $\epsilon$ &
  \begin{tabular}[c]{@{}c@{}}Trigger\\ Size\end{tabular} &
  \begin{tabular}[c]{@{}c@{}}Trigger\\ Location\end{tabular} &
  \begin{tabular}[c]{@{}c@{}}Trigger\\ Color\end{tabular} \\ \toprule
Gu et al.~\cite{badnets-evaluating-backdoor-attacks-on-dnns} &
  \begin{tabular}[c]{@{}c@{}}0.1\\ 0.3\\ 0.5\end{tabular} &
  \begin{tabular}[c]{@{}c@{}}Single pixel\\ Four pixels\end{tabular} &
  \begin{tabular}[c]{@{}c@{}}Bottom-right\\ Center\end{tabular} &
  \begin{tabular}[c]{@{}c@{}}White\\ Yellow\\ Patch\end{tabular} \\ \midrule
Salem et al.~\cite{salem2022dynamic} &
  \begin{tabular}[c]{@{}c@{}}0.1\\ 0.2\\ 0.3\\ 0.4\\ 0.5\end{tabular} &
  \begin{tabular}[c]{@{}c@{}}0.1\%\\ 0.5\%\\ 1.5\%\\ 2\%\\ 3.2\%\\ 4\%\end{tabular} &
  \begin{tabular}[c]{@{}c@{}}Corners\\ Top center\\ Bottom center\end{tabular} &
  \begin{tabular}[c]{@{}c@{}}Random\\ Dynamic\end{tabular} \\ \midrule
Liu et al.~\cite{liu2017trojaning} & 
  \begin{tabular}[c]{@{}c@{}} - \end{tabular} &
  \begin{tabular}[c]{@{}c@{}}4\%\\ 7\%\\ 10\%\end{tabular} &
  \begin{tabular}[c]{@{}c@{}} Bottom-right\end{tabular} &
  \begin{tabular}[c]{@{}c@{}} Dynamic \end{tabular} \\ \midrule
Kwon et al.~\cite{kwon2020multi} & 
  \begin{tabular}[c]{@{}c@{}} 0.1 \\ 0.25 \\ 0.5 \end{tabular} &
  \begin{tabular}[c]{@{}c@{}} 25\%\end{tabular} &
  \begin{tabular}[c]{@{}c@{}} Corners \end{tabular} &
  \begin{tabular}[c]{@{}c@{}} White \end{tabular} \\  \midrule
Tan et al.~\cite{bypassing-backdoor-detection-algorithms-in-deep-learning} & 
  \begin{tabular}[c]{@{}c@{}} 0.05 \end{tabular} &
  \begin{tabular}[c]{@{}c@{}} 8\%\end{tabular} &
  \begin{tabular}[c]{@{}c@{}} Bottom-right \end{tabular} &
  \begin{tabular}[c]{@{}c@{}} White \end{tabular} \\ \midrule
Feng et al.~\cite{feng2022stealthy} & 
  \begin{tabular}[c]{@{}c@{}} 0.005\\ 0.01\\ 0.02  \end{tabular} &
  \begin{tabular}[c]{@{}c@{}} - \end{tabular} &
  \begin{tabular}[c]{@{}c@{}} - \end{tabular} &
  \begin{tabular}[c]{@{}c@{}} Dynamic \end{tabular} \\ \midrule
Zhang et al.~\cite{zhang2021advdoor} &
  \begin{tabular}[c]{@{}c@{}} 0.1\\ 0.2\\ 0.3\\ 0.4\\ 0.5 \end{tabular} &
  \begin{tabular}[c]{@{}c@{}} - \end{tabular} &
  \begin{tabular}[c]{@{}c@{}} - \end{tabular} &
  \begin{tabular}[c]{@{}c@{}} Dynamic \end{tabular} \\ \midrule

Li et al.~\cite{li2021invisible} &
  \begin{tabular}[c]{@{}c@{}} 0.02\\ 0.04\\ 0.06\\ 0.08\\ 0.1 \end{tabular} &
  \begin{tabular}[c]{@{}c@{}} -  \end{tabular} &
  \begin{tabular}[c]{@{}c@{}} -  \end{tabular} &
  \begin{tabular}[c]{@{}c@{}} Dynamic  \end{tabular} \\ \midrule
  
Li et al.~\cite{invisible-backdoor-attacks-on-dnns-via-steganography-and-regularization} &
  \begin{tabular}[c]{@{}c@{}} 0.05\\ 0.1\\ 0.15\\ 0.2\\ 0.25 \end{tabular} &
  \begin{tabular}[c]{@{}c@{}} -  \end{tabular} &
  \begin{tabular}[c]{@{}c@{}} -  \end{tabular} &
  \begin{tabular}[c]{@{}c@{}} (Static)\\ imperceptible \\ backdoors \\ using \\ $L_{\rho}$ reqularization\\ \& steganography  \end{tabular} \\ \midrule
  
Nguyen et al.~\cite{wanet-imperceptible-warping-based-backdoor-attack} &
  \begin{tabular}[c]{@{}c@{}} 0.2 \end{tabular} &
  \begin{tabular}[c]{@{}c@{}} -  \end{tabular} &
  \begin{tabular}[c]{@{}c@{}} -  \end{tabular} &
  \begin{tabular}[c]{@{}c@{}} Dynamic  \end{tabular} \\ \midrule

Zeng et al.~\cite{zeng2021rethinking} &
  \begin{tabular}[c]{@{}c@{}} 0.1 \end{tabular} &
  \begin{tabular}[c]{@{}c@{}} -  \end{tabular} &
  \begin{tabular}[c]{@{}c@{}} -  \end{tabular} &
  \begin{tabular}[c]{@{}c@{}} Dynamic  \end{tabular} \\ \midrule

Barni et al.~\cite{barni2019new} &
  \begin{tabular}[c]{@{}c@{}} 0.2\\ 0.3\\ 0.4 \end{tabular} &
  \begin{tabular}[c]{@{}c@{}} -  \end{tabular} &
  \begin{tabular}[c]{@{}c@{}} -  \end{tabular} &
  \begin{tabular}[c]{@{}c@{}} Dynamic  \end{tabular} \\ \midrule
  
Chen et al.~\cite{chen2017targeted} &
  \begin{tabular}[c]{@{}c@{}} 0.02\\ 0.05\\ 0.1\\ 0.2 \end{tabular} &
  \begin{tabular}[c]{@{}c@{}} -  \end{tabular} &
  \begin{tabular}[c]{@{}c@{}} -  \end{tabular} &
  \begin{tabular}[c]{@{}c@{}} Static Pattern  \end{tabular} \\ \midrule

Nguyan et al.~\cite{nguyen2020input} &
  \begin{tabular}[c]{@{}c@{}} $\rho_{attack}=0.1$\\ $\rho_{cross-trigger}=0.1$ \end{tabular} &
  \begin{tabular}[c]{@{}c@{}} -  \end{tabular} &
  \begin{tabular}[c]{@{}c@{}} -  \end{tabular} &
  \begin{tabular}[c]{@{}c@{}} Dynamic trigger\\ crafted using a \\generator  \end{tabular} \\ \midrule

Doan et al.~\cite{doan2021backdoor} &
  \begin{tabular}[c]{@{}c@{}} $0.01$ \end{tabular} &
  \begin{tabular}[c]{@{}c@{}} -  \end{tabular} &
  \begin{tabular}[c]{@{}c@{}} -  \end{tabular} &
  \begin{tabular}[c]{@{}c@{}} (Dynamic)\\ imperceptible \\ noise crafted by \\minimizing \\wasserstein \\distance  \end{tabular} \\ \midrule

  Liu et al.~\cite{liu2020reflection} &
  \begin{tabular}[c]{@{}c@{}} 0.005\\ 0.009\\ 0.02\\ 0.03 \end{tabular} &
  \begin{tabular}[c]{@{}c@{}} -  \end{tabular} &
  \begin{tabular}[c]{@{}c@{}} -  \end{tabular} &
  \begin{tabular}[c]{@{}c@{}} (Static)\\ Reflection  \end{tabular} \\ \midrule

    Zhao et al.~\cite{zhao2022defeat} &
  \begin{tabular}[c]{@{}c@{}} 0.01\\ 0.05\\ 0.1\\ 0.15 \end{tabular} &
  \begin{tabular}[c]{@{}c@{}} -  \end{tabular} &
  \begin{tabular}[c]{@{}c@{}} -  \end{tabular} &
  \begin{tabular}[c]{@{}c@{}} (Dynamic)\\ imperceptible \\backdoor \\triggers  \end{tabular} \\ \midrule

  Saha et al.~\cite{saha2020hidden} &
  \begin{tabular}[c]{@{}c@{}} 0.125\\ 0.5 \end{tabular} &
  \begin{tabular}[c]{@{}c@{}} $1\%$\\ $6\%$  \end{tabular} &
  \begin{tabular}[c]{@{}c@{}} random location\\ right-corner(CIFAR10)  \end{tabular} &
  \begin{tabular}[c]{@{}c@{}} (Static)\\ random \\pixel \\pattern  \end{tabular} \\ \midrule

    Wang et al.~\cite{wang2022bppattack} &
  \begin{tabular}[c]{@{}c@{}} 0.025\\ 0.05\\ 0.1\\ 0.2\\ 0.3 \end{tabular} &
  \begin{tabular}[c]{@{}c@{}} -  \end{tabular} &
  \begin{tabular}[c]{@{}c@{}} -  \end{tabular} &
  \begin{tabular}[c]{@{}c@{}} (Dynamic)\\ imperceptible \\backdoor \\triggers  \end{tabular} \\ \toprule

\end{tabular}%
}
\end{table}

%% file: sections/3_threat_model.tex
\section{Threat Model}
\label{sec:threat_model}

We consider a \emph{gray-box} threat model as the attacker can freely modify a small portion of the training dataset and has no knowledge about the training algorithms or the models used by the victims. We also assume a \emph{dirty-label} backdoor attack meaning that the attacker can alter both the training samples and their labels. Even though this threat model is weaker than its counterpart (clean-label attack~\cite{clean-label-backdoor-attacks}), it is the most popular among the existing works~\cite{badnets-evaluating-backdoor-attacks-on-dnns,targeted-backdoor-attacks-on-deep-learning-systems-using-data-poisoning,liu2017trojaning,salem2022dynamic,deep-feature-space-trojan-attack-on-nns-by-controlled-detoxification,nguyen2020input}. Additionally, we target only transfer learning as it has become a very common practice as training from scratch can be very expensive and the weights of state-of-the-art models like VGG and ResNet trained on ImageNet are publicly available~\cite{badnets-evaluating-backdoor-attacks-on-dnns}. This threat model is realistic as large datasets like ImageNet~\cite{imagenet-a-large-scale-hierarchical-image-database} are crowdsourced from untrusted sources, and malicious data can evade human inspection~\cite{large-dataset-pyrrhic-win}.

We consider the following metrics:
\begin{compactenum}
    \item \textbf{Attack Success Rate} (ASR): measures the backdoor performance of the model on a fully poisoned dataset $D_{poison}$, i.e., $\epsilon = 1$. 
    It can be computed by $ASR = \frac{\sum_{i=1}^{N}\mathbb{I}(F_{\hat{\theta}}(\hat{x_i})=y_t)}{N}$ where $F_{\hat{\theta}}$ is the poisoned model, $\hat{x_i}$ is a poisoned input, $\hat{x_i} \in D_{poison}$, $y_t$ is the target class, and $\mathbb{I}(x)$ is a function that returns $1$ if $x$ is true and $0$ otherwise. 
    \item \textbf{Clean Accuracy Drop} (CAD): measures the effect of the backdoor attack on the original task. It is calculated by comparing the performance of the poisoned and clean models on a clean holdout validation set $D_{valid} $, i.e., $\epsilon = 0$. The accuracy drop should be small to keep the attack stealthy. 
\end{compactenum}

%% file: sections/4_experiments.tex
\section{Experiments}
\label{sec:experiments}


We systematically evaluate the backdoor attacks on 4 DNN models and four poisoning rates for all three attacks. WaNet and SSBA were tested in CIFAR10 and TinyImagenet, while BadNets was also tested in MNIST. In BadNets, we also use 3 trigger sizes, 5 trigger positions, and 3 trigger colors. We also train a clean model for each dataset and architecture ($3\times 4$). Each of these experiments was repeated 5 times. Thus, we train 11,120 backdoored models and 60 clean models in total.

\subsection{Experimental Matrix}


\textbf{Datasets.} 
We evaluated our approach using MNIST, CIFAR10, and TinyImageNet.

\begin{compactitem}
    \item MNIST~\cite{lecun1998mnist} is a dataset of 70,000 grayscale images of handwritten digits, each $28\times28$ pixels in size and belonging to one of 10 different classes. We converted the images to RGB format for our evaluation and resized them to $64\times64$ pixels~\footnote{Data resizing is used to adapt inputs to the chosen networks, which require a minimum input size. Additionally, we experiment with different input sizes to better generalize the results.}.
    \item CIFAR10~\cite{Krizhevsky09learningmultiple} is a dataset of 60,000 RGB images, each $32\times32$ pixels in size and belonging to one of 10 different classes, with 6,000 images per class. Similar to MNIST, we resized the images to $128\times128$ pixels for compatibility.
    \item TinyImageNet~\cite{le2015tiny} is a dataset of 120,000 RGB images belonging to 200 different classes, each $64\times64$ pixels in size. We also resized these images to $224\times224$ pixels.
\end{compactitem}

\textbf{Model Architectures.} 
In our experiments, we selected four standard benchmark DNNs for evaluation: AlexNet, GoogLeNet, VGG-19\textunderscore BN, and ResNet-152. To utilize transfer learning and extract features from these models, we froze the parameters for all layers (except for the last fully connected layer and the batch-normalization layers for ResNet, VGG, and GoogLeNet). This allowed us to leverage the pre-trained models while focusing on the task of interest. AlexNet, however, resists backdoor injection when being operated by transfer learning (i.e., we reach low ASR when freezing all layers except the last one). Thus, for a more suitable analysis, our transfer learning setup for AlexNet is to freeze the layers up to its classifier's module (more in~\cref{sec:effect_of_model_architecture}).

These DNNs were selected based on their demonstrated performance on various tasks and widespread use as benchmarks in the field. AlexNet, a CNN introduced by Krizhevsky et al.~\cite{krizhevsky2017imagenet}, was the first successful CNN to demonstrate superior performance on the ImageNet dataset~\cite{fei2009imagenet}. In PyTorch implementation for AlexNet (which we use for our study in this work), it consists of two main modules: the features module, which itself consists of convolutions and pooling layers, and the classifier module, which is composed of fully connected layers for the final classification task. GoogLeNet, introduced by Szegedy et al.~\cite{szegedy2015going}, is a variant of the Inception architecture that won the 2014 ImageNet Large Scale Visual Recognition Challenge. VGG-19\textunderscore BN is a variant of the VGG network~\cite{simonyan2014very} that incorporates batch normalization~\cite{ioffe2015batch} and has achieved strong performance on a range of tasks. Finally, ResNet-152 is a residual network~\cite{he2016deep} with 152 layers that also achieved state-of-the-art performance on several tasks. By using these well-established DNNs, we ensure the reliability and generalizability of our results.

\textbf{Trigger Characteristics.}
The following trigger characteristics apply to the BadNets attack:

$\triangleright$ Trigger Size:
When using the BadNets attack, we focus on the square trigger pattern, a commonly used trigger in backdoor attacks on image classification tasks~\cite{badnets-evaluating-backdoor-attacks-on-dnns, systematic-evaluation-of-backdoor-data-poisoning-attacks-on-image-classifiers}. This trigger consists of a square patch injected into the training images and used to manipulate the model's behavior. In~\cite{systematic-evaluation-of-backdoor-data-poisoning-attacks-on-image-classifiers}, the square trigger proved the most effective, so we did not consider blending overlay triggers. To evaluate the effectiveness of the attack under different conditions, we varied the width and height of the trigger as a percentage of the width and height of the sampled training image, using values of 4\%, 6\%, and 8\%, which allowed us to assess the trigger size's impact on the attack's performance. These trigger sizes cover most of the trigger sizes considered in the literature while being realistic.

\begin{figure}[!ht]
     \centering
     \begin{subfigure}{0.08\textwidth}
         \centering
         \includegraphics[width=1\textwidth]{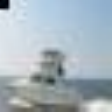}
         \caption{}
         \label{fig:trig_pos_top_left}
     \end{subfigure}
     \hfill
     \begin{subfigure}{0.08\textwidth}
         \centering
         \includegraphics[width=\textwidth]{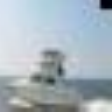}
         \caption{}
         \label{fig:trig_pos_top_right}
     \end{subfigure}
     \hfill
     \begin{subfigure}{0.08\textwidth}
         \centering
         \includegraphics[width=\textwidth]{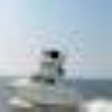}
         \caption{}
         \label{fig:trig_pos_middle}
     \end{subfigure}
     \hfill
     \begin{subfigure}{0.08\textwidth}
         \centering
         \includegraphics[width=\textwidth]{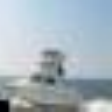}
         \caption{}
         \label{fig:trig_pos_bottom_left}
     \end{subfigure}
     \hfill
     \begin{subfigure}{0.08\textwidth}
         \centering
         \includegraphics[width=\textwidth]{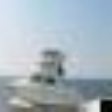}
         \caption{}
         \label{fig:trig_pos_bottom_right}
     \end{subfigure}
        \caption{Trigger patterns with different trigger positions (top-left, top-right, middle, bottom-left, bottom-right) applied to an image from the CIFAR10 dataset.}
        \label{fig:asr_graph_classification}
\end{figure}

$\triangleright$ Trigger Position:
We inject the square trigger into five locations in the poisoned images: the top-left, top-right, middle, bottom-left, and bottom-right positions. This allowed us to evaluate the impact of the trigger position on the performance of the attack and to identify any trends or patterns that may be present. \autoref{fig:asr_graph_classification} illustrates an example image from the CIFAR10 dataset with triggers embedded at various positions. By studying the attack under these different conditions, we could gain a deeper understanding of the factors that influence the success of a backdoor attack and develop more effective defense strategies.

$\triangleright$ Trigger Color:
The color of the trigger pattern is another critical factor to consider in the design of a backdoor attack. In our experiments, we evaluated the performance of the attack using three different trigger colors: black, white, and green. The green trigger was randomly picked to avoid biases that extreme values like black (0, 0, 0) or white (255, 255, 255) may create. To this end, we used Python's pseudorandom generator to retrieve three RGB values (one for each channel). The values are (102, 179, 92). The MNIST dataset only has one channel, so we run experiments of green color on CIFAR10 and TinyImageNet datasets. By comparing the results obtained with these different trigger colors, we could understand how the color of the trigger affects the effectiveness of the attack and identify any trends or patterns that may be present. This information is valuable for understanding these attacks' behavior and developing more effective defense strategies.

\textbf{Poisoning Rate.}
\label{sec:poisoning-rate}
One key factor that impacts the backdoor's effectiveness is the poisoning rate, which refers to the percentage of training images injected with the backdoor trigger. This parameter is not limited to the BadNets attack but is important in all backdoor attacks. In our experiments, we replaced clean images with their poisoned counterparts to avoid altering the number of training samples in the dataset. We also varied the poisoning rate to study its impact on attack performance. This allowed us to study the effect of different poisoning rates on the attacks' success. Additionally, we chose low poisoning rates because the backdoor should affect the original task as little as possible, and given the amount of data that modern deep learning systems need, large poisoning rates can be unrealistic~\cite{targeted-backdoor-attacks-on-deep-learning-systems-using-data-poisoning}. Thus, we defined four values: 0.5\%, 1\%, 1.5\%, and 2\%.

\subsection{Experimental Setup}


\textbf{Training Procedure.}
We chose the Adam algorithm and cross-entropy loss as an optimizer and the criterion in our experiments. However, in one setting (TinyImageNet + VGG), the Adam optimizer yielded poor performance (around 37\%), and we had to use SGD, which resulted in an accuracy of around 72\%. Furthermore, we experimentally set the learning rate to 0.001 and the number of epochs to 20, where we achieve training convergence and good generalization in the test set.
Each dataset's batch size is different to fit into the GPU's memory. For the small datasets (MNIST and CIFAR10), the batch size is 128, and for TinyImageNet is 32. 
Each experiment was repeated five times to reduce the effects of randomness caused by stochastic gradient descent and initialization.

\subsection{Results and Analysis}


\subsubsection{Clean Accuracy Drop}
\label{sec:cad}

The backdoor should remain stealthy in the deployed model to avoid raising any suspicions. Thus, the model's performance on the original task should not be affected by the backdoor insertion. To ensure this is true in our experiments, we calculate the arithmetic mean of the accuracy of all the clean models we trained ($\epsilon = 0$) and compared it to the mean of the accuracy of the poisoned models. We show the results in~\autoref{tab:cad}, where we use bold for the value with the largest difference from the clean model. We see that the difference introduced by the backdoor is really small. In almost all cases, the accuracy is decreased but less than 1\%. We see only in one case (TinyImagenet + ResNet) a performance drop of around 2\%. ResNet is the best-performing model with TinyImageNet (clean accuracy 83.96\%), and even a small change in the training data affects the model's generalization.
\textbf{Additionally, the performance drop is positively correlated with $\epsilon$ and as $\epsilon$ gets larger, the drop is increased as well}.

From~\autoref{tab:cad}, we can also see that our models perform well for the datasets tested. However, AlexNet is not very accurate with TinyImageNet and has an accuracy of 21.73\%($\pm0.6067$). As we discussed in~\Cref{sec:effect_of_model_architecture}, the backdoor did not work in this case if we freeze all the layers up to the fully connected layers. Thus, we had to unfreeze a few layers from the feature extractor for the backdoor to be more effective. This resulted in lower performance for the original task as we altered the weights of the feature extractor. If we keep these layers frozen, the model's accuracy for clean inputs is around 46\%. \textbf{Thus, we conclude that the backdoor attack is more effective if the model is trained from scratch or has a large number of trainable layers}.


\begin{table*}[htb]
\centering
\caption{Clean accuracy comparison between clean and poisoned models. We show in bold the settings that have the largest difference with the clean model's ($\epsilon = 0$) performance.}
\resizebox{0.7\textwidth}{!}{%
\label{tab:cad}
\begin{tabular}{c|c|ccccc}
\toprule
  \multirow{2}{*}{Dataset} &
  \multirow{2}{*}{Model} &
  \multicolumn{5}{c}{$\epsilon$ (\%)} \\
 &
   &
  \multicolumn{1}{c|}{0} &
  \multicolumn{1}{c|}{0.5} &
  \multicolumn{1}{c|}{1} &
  \multicolumn{1}{c|}{1.5} &
  2 \\ \hline
\multirow{4}{*}{MNIST} &
  AlexNet &
  \multicolumn{1}{c|}{98.50$\pm$0.1915} &
  \multicolumn{1}{c|}{98.41$\pm$0.1883} &
  \multicolumn{1}{c|}{98.37$\pm$0.1876} &
  \multicolumn{1}{c|}{\textbf{98.30$\pm$0.2468}} &
  98.31$\pm$0.2136 \\ 
 &
  GoogLeNet &
  \multicolumn{1}{c|}{98.75$\pm$0.1191} &
  \multicolumn{1}{c|}{98.67$\pm$0.1363} &
  \multicolumn{1}{c|}{98.64$\pm$0.1654} &
  \multicolumn{1}{c|}{98.62$\pm$0.1817} &
  \textbf{98.58$\pm$0.2173} \\ 
 &
  ResNet &
  \multicolumn{1}{c|}{98.83$\pm$0.1846} &
  \multicolumn{1}{c|}{98.64$\pm$0.3198} &
  \multicolumn{1}{c|}{98.50$\pm$0.4217} &
  \multicolumn{1}{c|}{98.33$\pm$0.4882} &
  \textbf{98.19$\pm$0.6144} \\ 
 &
  VGG &
  \multicolumn{1}{c|}{99.09$\pm$1.1671} &
  \multicolumn{1}{c|}{99.22$\pm$1.1597} &
  \multicolumn{1}{c|}{\textbf{99.34$\pm$0.1769}} &
  \multicolumn{1}{c|}{99.31$\pm$0.4784} &
  99.30$\pm$0.2782 \\ \hline
\multirow{4}{*}{CIFAR10} &
  AlexNet &
  \multicolumn{1}{c|}{85.17$\pm$0.3677} &
  \multicolumn{1}{c|}{84.89$\pm$0.4034} &
  \multicolumn{1}{c|}{84.68$\pm$0.4217} &
  \multicolumn{1}{c|}{84.52$\pm$0.4050} &
  \textbf{84.40$\pm$0.4397} \\ 
 &
  GoogLeNet &
  \multicolumn{1}{c|}{92.54$\pm$0.1464} &
  \multicolumn{1}{c|}{92.38$\pm$0.2023} &
  \multicolumn{1}{c|}{92.33$\pm$0.2190} &
  \multicolumn{1}{c|}{92.22$\pm$0.2172} &
  \textbf{92.18$\pm$0.1961} \\ 
 &
  ResNet &
  \multicolumn{1}{c|}{96.88$\pm$0.1449} &
  \multicolumn{1}{c|}{96.68$\pm$0.1983} &
  \multicolumn{1}{c|}{96.61$\pm$0.2675} &
  \multicolumn{1}{c|}{96.58$\pm$0.3037} &
  \textbf{96.56$\pm$0.3779} \\ 
 &
  VGG &
  \multicolumn{1}{c|}{93.02$\pm$0.4733} &
  \multicolumn{1}{c|}{92.87$\pm$0.5260} &
  \multicolumn{1}{c|}{92.90$\pm$0.4743} &
  \multicolumn{1}{c|}{92.85$\pm$0.4712} &
  \textbf{92.77$\pm$0.5308} \\ \hline
\multirow{4}{*}{TinyImageNet} &
  AlexNet &
  \multicolumn{1}{c|}{21.73$\pm$0.6067} &
  \multicolumn{1}{c|}{21.60$\pm$0.6881} &
  \multicolumn{1}{c|}{21.49$\pm$0.7208} &
  \multicolumn{1}{c|}{21.17$\pm$0.7209} &
  \textbf{20.89$\pm$0.7470} \\ 
 &
  GoogLeNet &
  \multicolumn{1}{c|}{70.07$\pm$0.1688} &
  \multicolumn{1}{c|}{70.02$\pm$0.2322} &
  \multicolumn{1}{c|}{69.96$\pm$0.2569} &
  \multicolumn{1}{c|}{69.86$\pm$0.2513} &
  \textbf{69.79$\pm$0.2574} \\ 
 &
  ResNet &
  \multicolumn{1}{c|}{83.96$\pm$0.1927} &
  \multicolumn{1}{c|}{82.90$\pm$0.5713} &
  \multicolumn{1}{c|}{82.45$\pm$0.7705} &
  \multicolumn{1}{c|}{82.09$\pm$0.9795} &
  \textbf{81.90$\pm$0.9917} \\ 
 &
  VGG &
  \multicolumn{1}{c|}{72.66$\pm$0.2265} &
  \multicolumn{1}{c|}{72.60$\pm$0.2211} &
  \multicolumn{1}{c|}{72.51$\pm$0.2564} &
  \multicolumn{1}{c|}{72.41$\pm$0.2315} &
  \textbf{72.33$\pm$0.2301} \\ \bottomrule
\end{tabular}
}
\end{table*}

\subsubsection{Effect of Model Architecture}
\label{sec:effect_of_model_architecture}

In most cases, VGG is the least robust to backdoor attacks, especially on the CIFAR10 dataset, while AlexNet is the most robust to poisoning on all three datasets. For example, in~\autoref{fig:rate-vs-size-black-bottom-right}, the attack success rate of VGG is always higher than other models. 
Specifically, the attack performance of VGG with a small poisoning rate, i.e., 0.005, is higher than the other models. VGG has the most neurons among the four models, leading to a larger capacity to learn the backdoor functionality. 
With the increasing poisoning rate, the ASR of the ResNet and GoogLeNet increases and is similar to VGG's. \textbf{We believe that models with larger capacities are more vulnerable to backdoors as they can encode more patterns in their weights even from a very small part of the dataset}.

\begin{figure}
    \centering
    \includegraphics[width=0.95\linewidth]
    {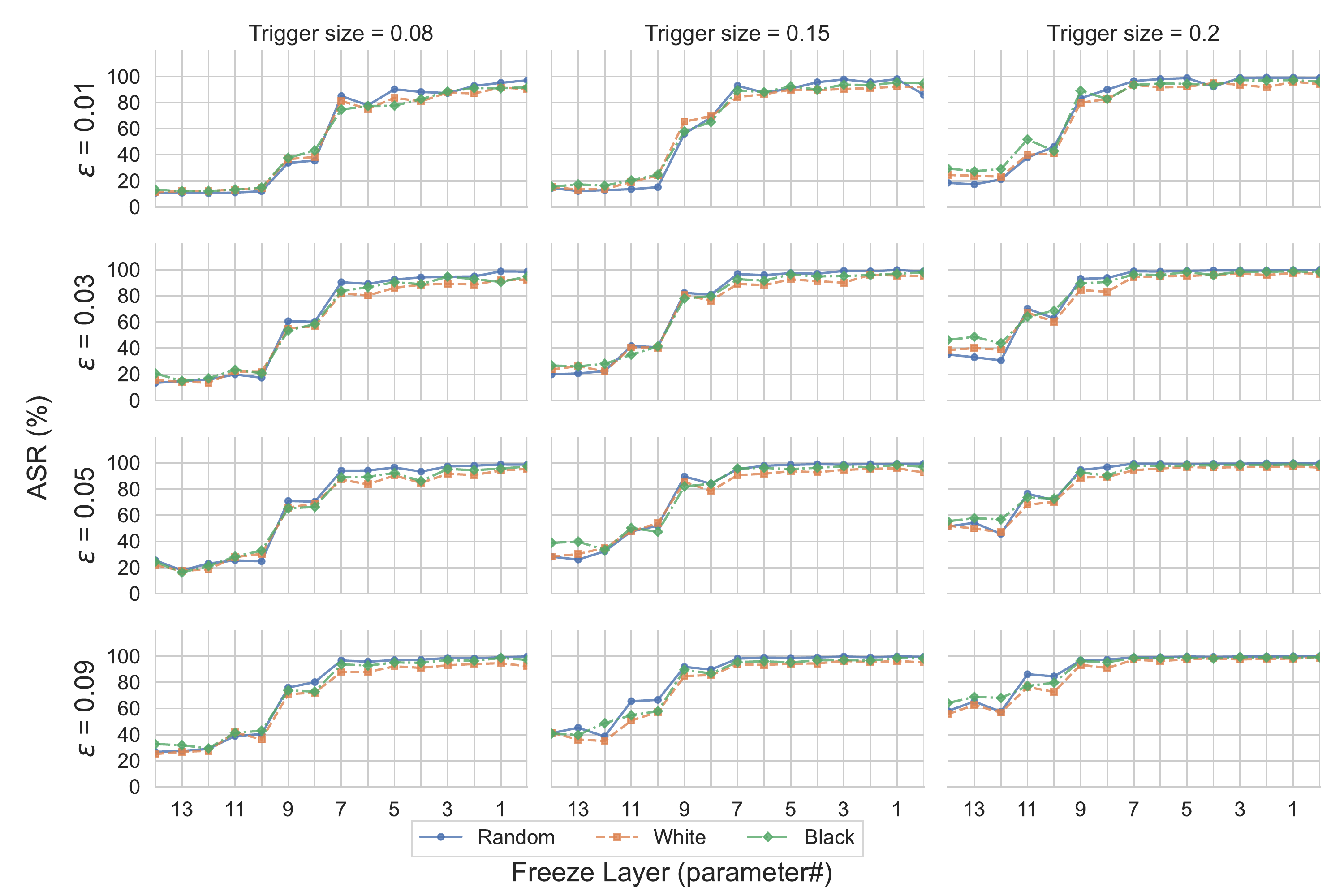}
    \caption{AlexNet on CIFAR10: FreezeLayer effect vs. size and rate, trigger at bottom-right}
    \label{fig:alexnet_cifar10_freezelayer_sizerate_buttomright}
\end{figure}

Additionally, if we freeze the feature extractor layers in AlexNet like the other three models, the ASR is nearly 0\%. Because of this, we decided to unfreeze AlexNet parameters layer by layer (from 14 to 0) to see from which layer it starts to react positively on the injected backdoor. Appendix \autoref{fig:alexnet_mnist_freezelayer_sizerate_buttomright} and \autoref{fig:alexnet_cifar10_freezelayer_sizerate_buttomright} show the results of our experiments on MNIST and CIFAR10. When we unfreeze the classifier module completely, the network starts to learn the backdoor. This can be observed on the plots when the network is unfrozen up to the 7th parameter. Thereafter, there is a surge in most of the plots from this point, showing that the backdoor has started to work. After this experiment, we decided to do the experiments with AlexNet by retraining the whole classifier module and freezing the feature Module. Nonetheless, the results show, except for trigger positions in the middle for CIFAR10 (\autoref{fig:rate-vs-size-white-middle} and~\autoref{fig:rate-vs-size-black-middle}), in all other experiments, the backdoor attack fails to reach high ASR on AlexNet. Interestingly, the classifier modules in AlexNet and VGG are very similar (both having 3 fully connected layers with 4,096 neurons in each layer). The main difference between these two is that in AlexNet, the two dropout layers precede the linear layers, but in VGG, they succeed (this means that in AlexNet, the first dropout will affect the last convolution layer in the features module, while in VGG both dropouts will affect the former fully connected neurons before them). VGG is a deeper and more complicated network than AlexNet, making it more vulnerable to backdoor triggers. However, the reason for AlexNet's robustness against backdoors is not merely its smaller capacity because unfreezing the classifier part improves backdoor learning. From~\cite{salem2020don}, we know that dropouts can affect the learning process of a network and cause a network to learn a deliberate backdoor. We correspondingly assume that the role of dropout layers and their inactivity during test time may affect the backdoor success. Nonetheless, we are not 100\% sure about this, and more experimental studies are needed to be done in the future to uncover the primary reason.

\textbf{AlexNet has demonstrated to be a very robust network on simple square shape backdoor patterns compared to the other three benchmark networks. It seems that the most important parameter which could affect the ASR on AlexNet is the trigger size.} \autoref{fig:AlexNet_fm_cifar10} displays the output of the AlexNet feature module on the same poisoned image with different trigger sizes (the dissimilarity of activations based on trigger size can be observed by comparing two feature map differences on the right).

\subsubsection{Effect of the Trigger Size}

In our experiments, we see that by only changing the trigger size, we can create very effective backdoors. For example, in~\autoref{fig:rate-vs-size-white-bottom-left}, we see that the ASR for AlexNet and MNIST is very low (around 10\%) in all cases when the trigger size is 4\% or 6\%. However, in the same setting, changing the trigger size to 8\% could lead to an ASR as high as 80\%. Similar behavior is shown in~\autoref{fig:rate-vs-size-black-middle} and in~\autoref{fig:rate-vs-size-white-middle} for CIFAR10 and all the models.

\begin{figure}                        
    \centering
    \includegraphics[width=0.85\linewidth]{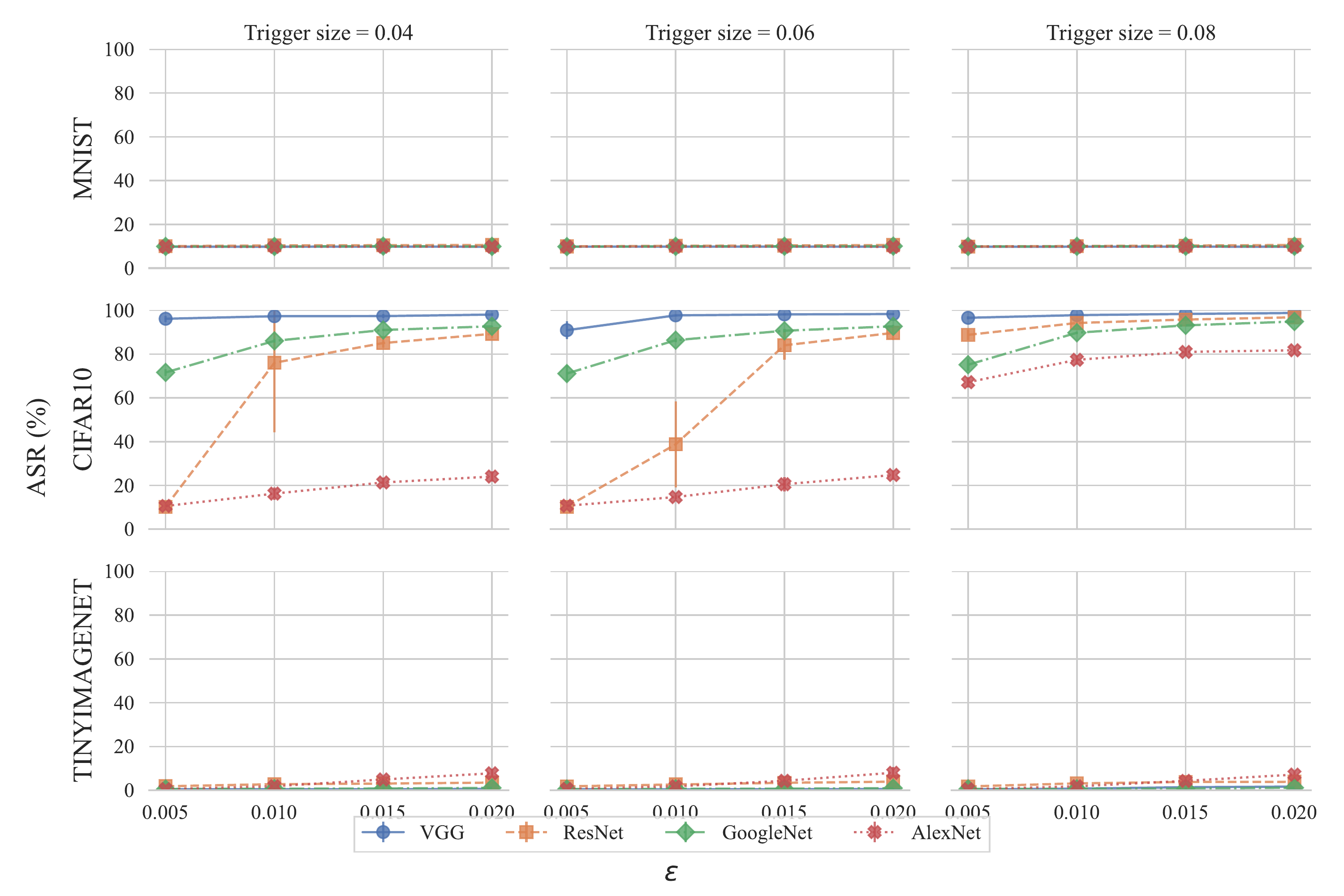}
    \caption{Rate vs. size, black color, trigger at bottom-right}
    \label{fig:rate-vs-size-black-bottom-right}
\end{figure}

For the CIFAR10 dataset, we see that the trigger size is the most influential for AlexNet. The feature extractor of the model is unfrozen in AlexNet, so the model can learn easier to spot larger triggers. However, there are multiple cases where changing the trigger size leads to high ASR for the other models. For example, in~\autoref{fig:rate-vs-size-black-bottom-right}, we see that the ASR for ResNet increases from 40\% to more than 90\% when $\epsilon = 0.01$ and the size is increased from 6\% to 8\%. Similarly, 
in~\autoref{fig:rate-vs-size-black-top-right}, the ASR for GoogLeNet is increased from around 10\% to more than 80\%.

When we insert the trigger in the middle, for the MNIST and CIFAR10 datasets, the ASR of all models (except AlexNet) is around 10\% with a trigger size less than 8\%. However, it increases significantly with a trigger size of 8\%. 
In TinyImageNet, the ASR is low when the trigger is not placed in the middle of the image. However, even in these cases, increasing the size may increase the ASR (
\autoref{fig:rate-vs-size-green-top-left}). \textbf{Thus, we conclude that the trigger size can significantly affect the ASR}. 


\subsubsection{Effect of the Trigger Position}

For the MNIST and CIFAR10 datasets, with a trigger size is less than 8\%, there is no noticeable difference in the ASR when it is injected in the corners. However, there is a decrease in the ASR for all models when the trigger is injected in the middle. 
With trigger size increasing to 8\%, the trigger position has an unnoticeable impact on the attack performance. 
On the contrary, for the TinyImageNet dataset, all models are robust to the backdoor attack when the trigger is not injected in the middle. With the trigger in the middle, there is a significant rise in the ASR for all models. 
This could explain that, in general, images in TinyImageNet are not centered, in contrast with those in MNIST and CIFAR10. Therefore, for TinyImageNet, triggers placed in the middle can achieve high ASR without a noticeable degradation on the main task. However, the model cannot recognize triggers placed in the corners or are small, i.e., less than 8\%.

\textbf{All these show that in our experiments, no position universally leads to a more successful backdoor attack. The most effective position is different for every dataset and depends on the dataset's properties and the way the models learn}. 

\subsubsection{Effect of the Trigger Color}

\begin{figure}
    \centering
    \includegraphics[width=0.85\linewidth]{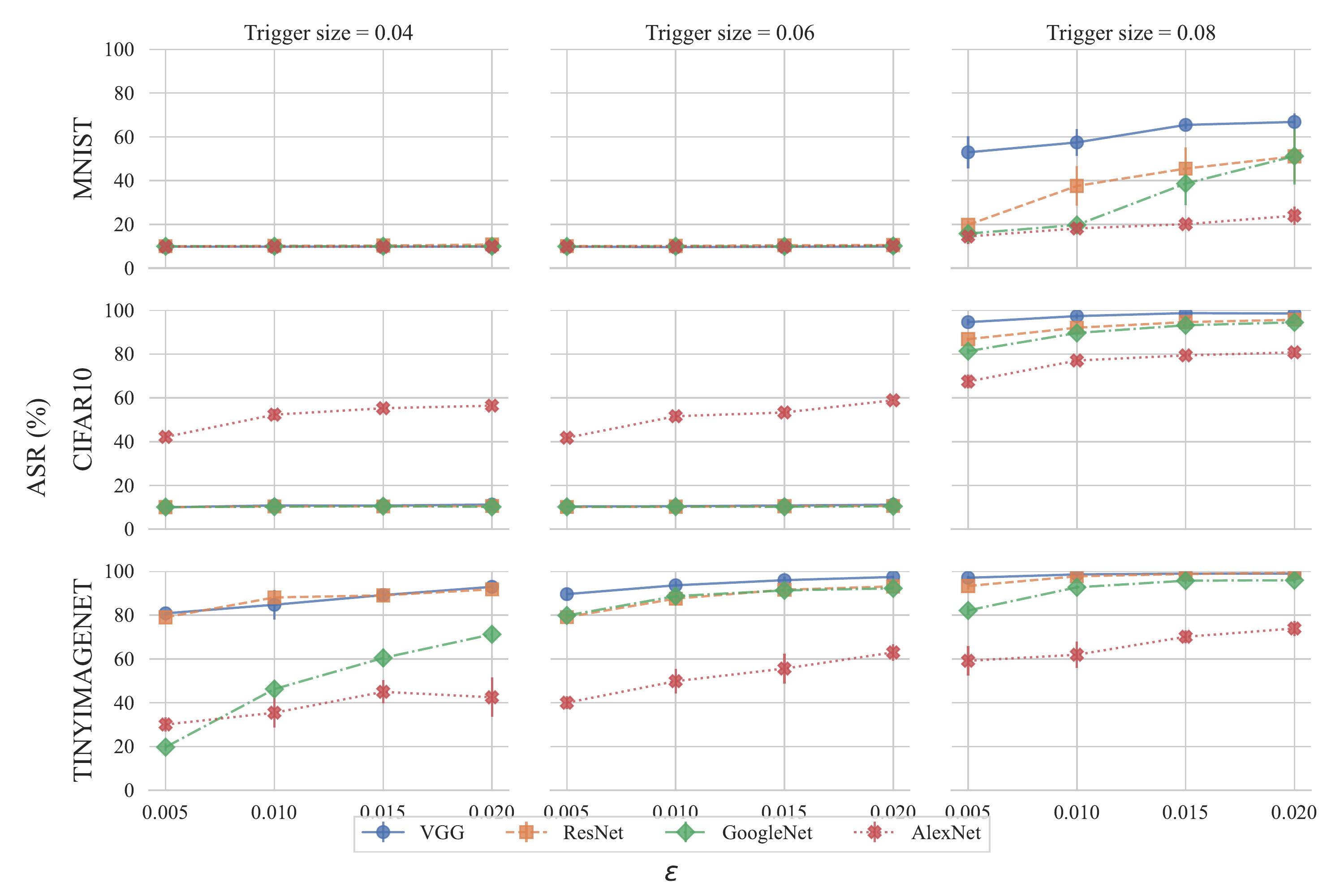}
    \caption{Rate vs. size, black color, trigger at middle}
    \label{fig:rate-vs-size-black-middle}
\end{figure}

\begin{figure}
    \centering
    \includegraphics[width=0.85\linewidth]{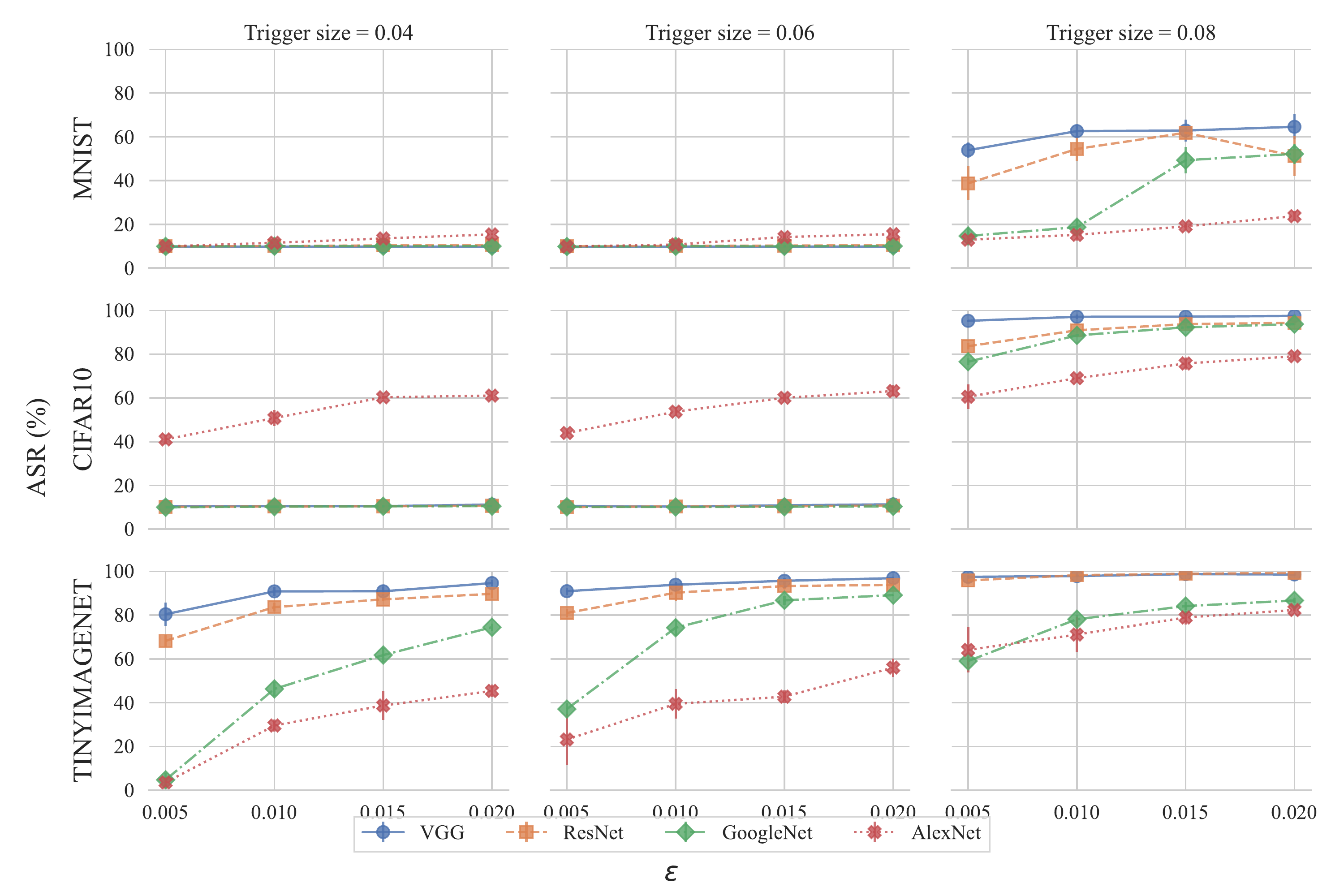}
    \caption{Rate vs. size, white color, trigger at middle}
    \label{fig:rate-vs-size-white-middle}
\end{figure}

For MNIST, the ASR is low for black triggers placed in the corners. The effect is expected as the training images in MNIST contain many black pixels by default, and the model cannot identify our black trigger as a feature. However, white triggers placed in the corners are effective due to their contrast with the black background. For both colors (black and white), the trigger should be large (8\%) to start having an effect on the ASR when placed in the middle (\autoref{fig:rate-vs-size-black-middle} and~\autoref{fig:rate-vs-size-white-middle}) as it overlaps the sample's main information, i.e., the number.

\begin{figure}
    \centering
    \includegraphics[width=0.85\linewidth]{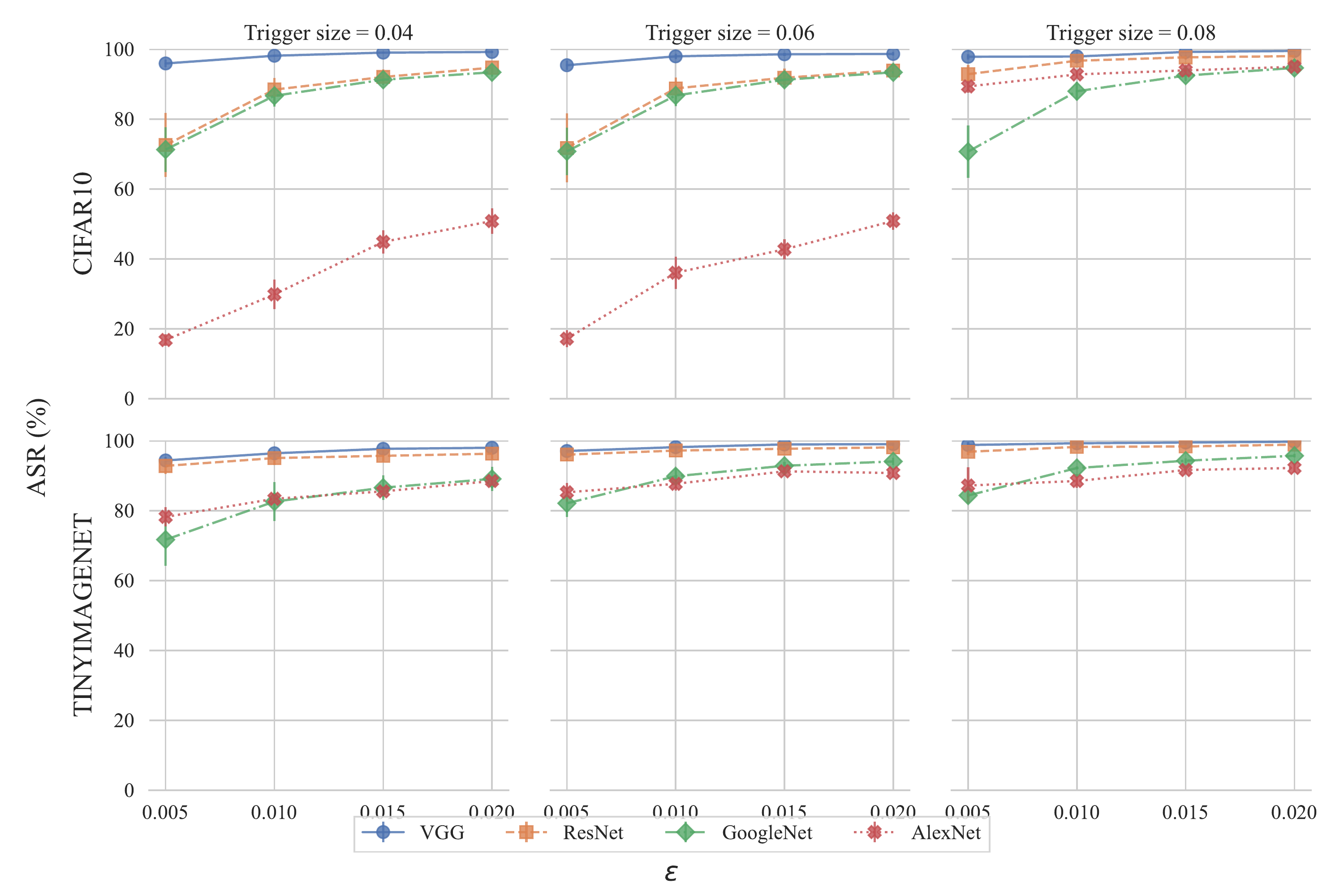}
    \caption{Rate vs. size, green color, trigger at middle}
    \label{fig:rate-vs-size-green-middle}
\end{figure}

\begin{figure}
    \centering
    \includegraphics[width=0.85\linewidth]{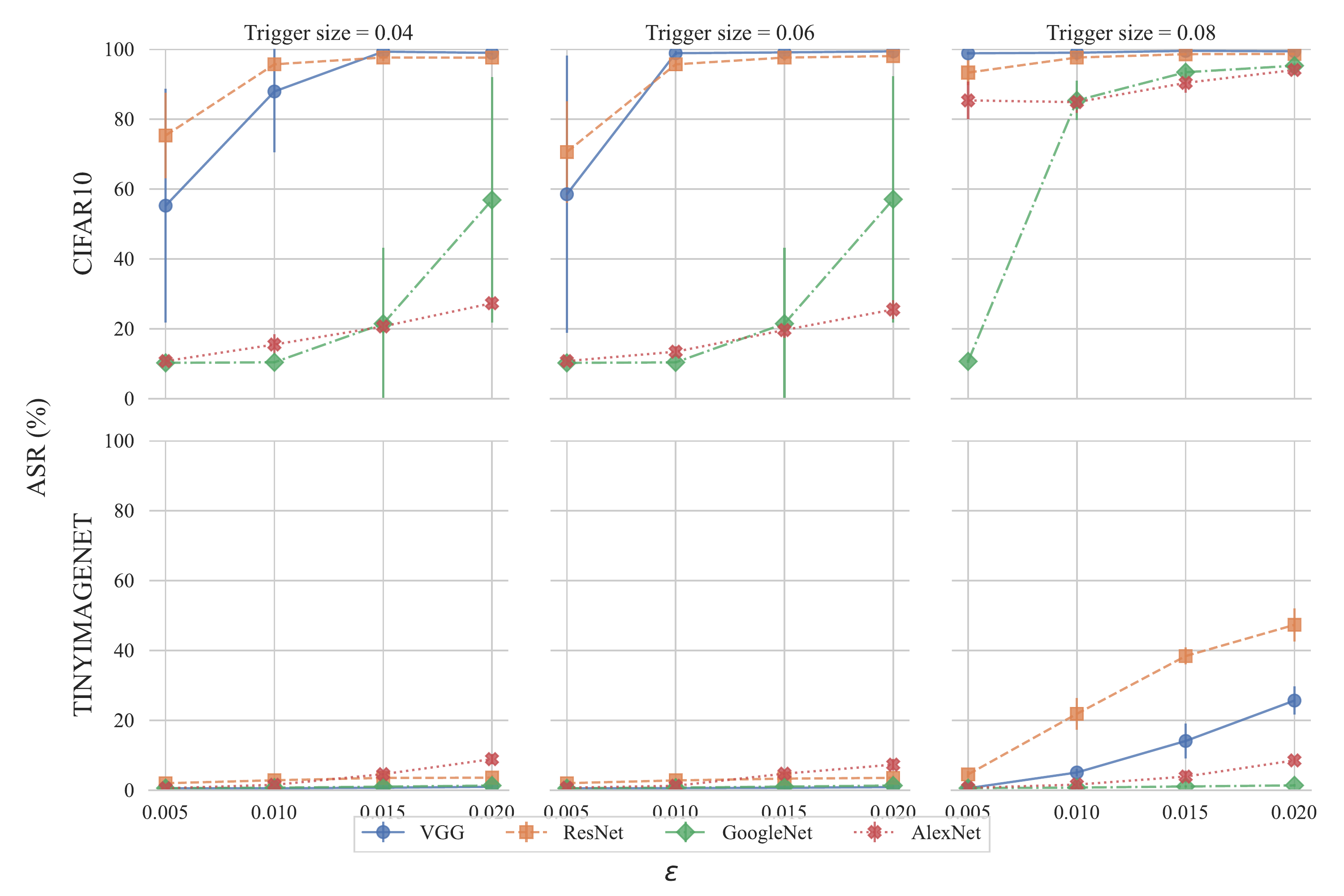}
    \caption{Rate vs. size, green color, trigger at top-left}
    \label{fig:rate-vs-size-green-top-left}
\end{figure}

In TinyImageNet, when the trigger is placed in the corners, in most cases, the ASR is around 0\%. However, when the trigger has a size of 8\% and is green 
, 
the ASR can be increased up to 40\% (\autoref{fig:rate-vs-size-green-top-left}). Additionally, when the trigger is inserted in the middle, the backdoor works in all cases but is more effective when green (\autoref{fig:rate-vs-size-green-middle}).

\begin{figure}
    \centering
    \includegraphics[width=0.85\linewidth]{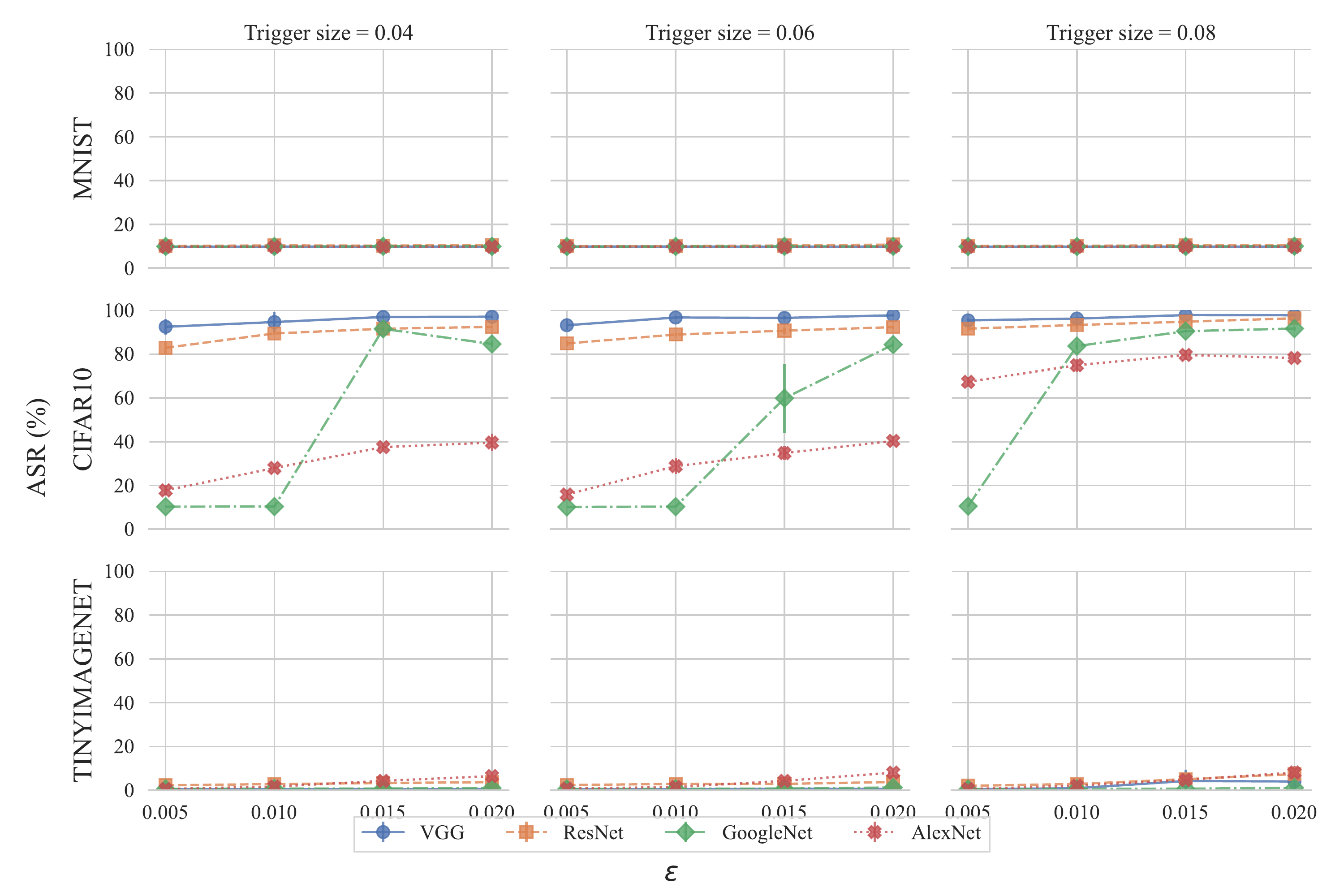}
    \caption{Rate vs. size, black color, trigger at top-right}
    \label{fig:rate-vs-size-black-top-right}
\end{figure}

\begin{figure}
    \centering
    \includegraphics[width=0.85\linewidth]{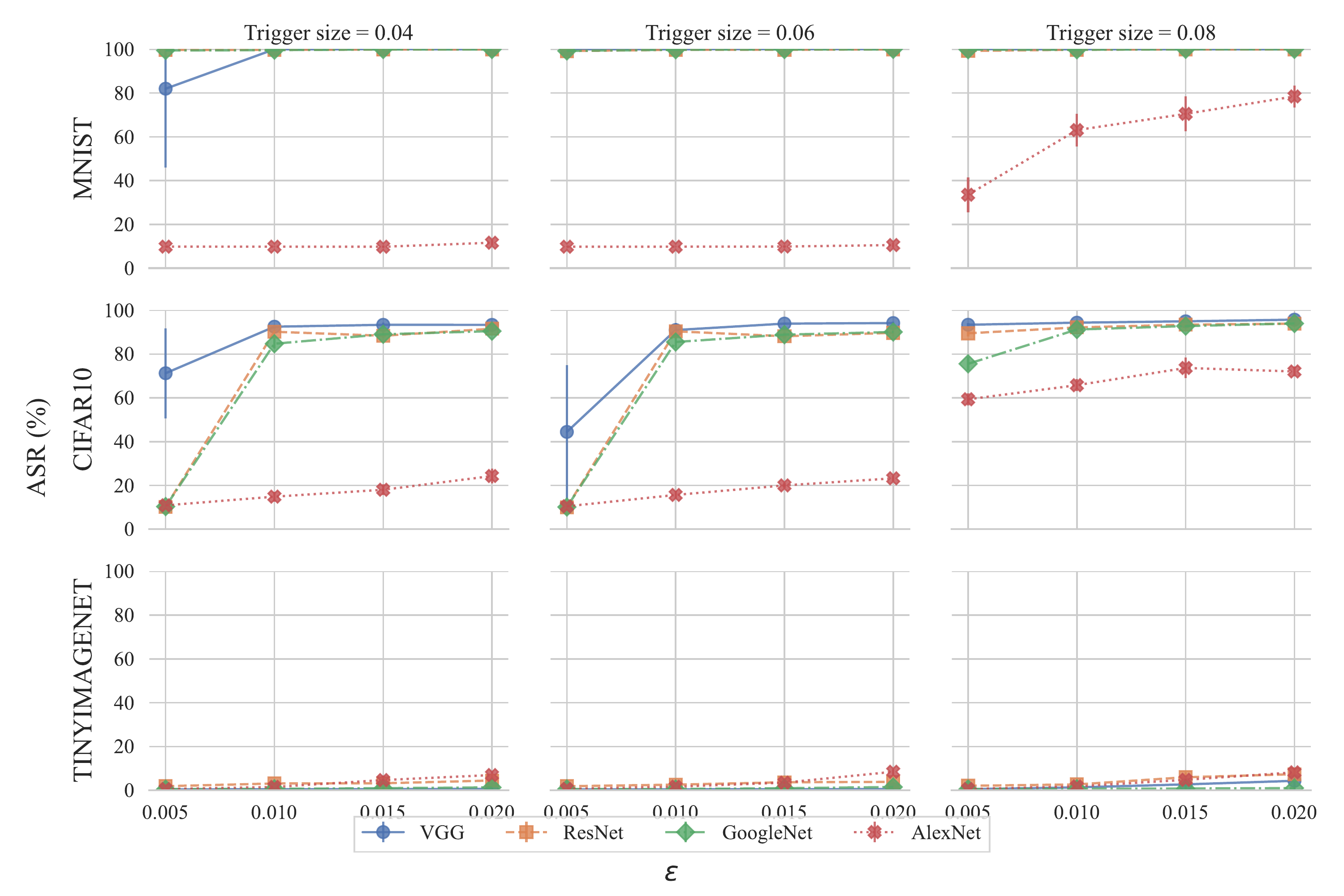}
    \caption{Rate vs. size, white color, trigger at bottom-left}
    \label{fig:rate-vs-size-white-bottom-left}
\end{figure}

\begin{figure}
    \centering
    \includegraphics[width=0.85\linewidth]{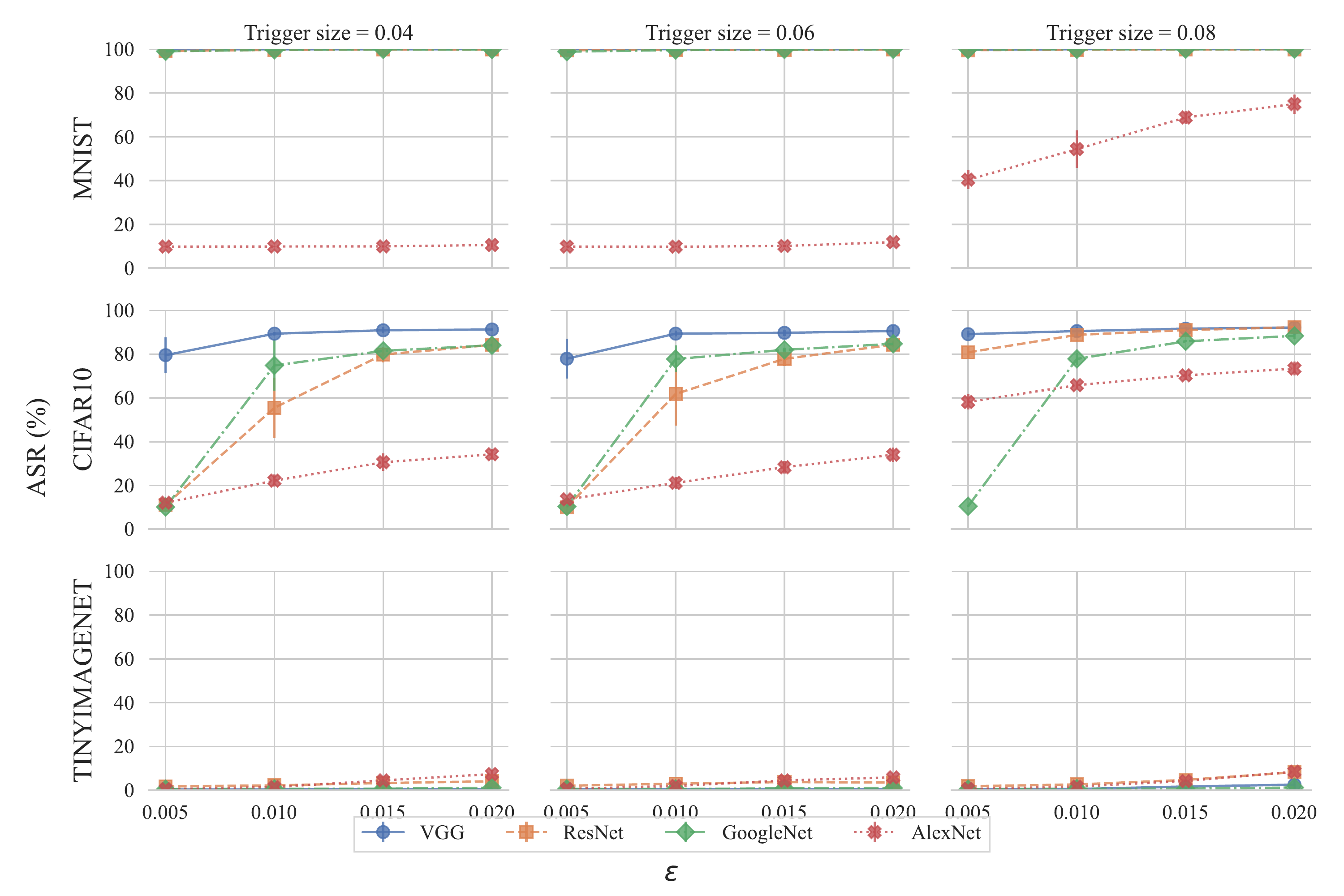}
    \caption{Rate vs. size, white color, trigger at top-right}
    \label{fig:rate-vs-size-white-top-right}
\end{figure}

In CIFAR10, we see that in some cases for small triggers ($<$ 8\%), GoogLeNet is more effective with white triggers. For example, comparing~\autoref{fig:rate-vs-size-black-top-right} and~\autoref{fig:rate-vs-size-white-bottom-left} we see that for trigger size 4\% and $\epsilon$ = 0.5\%, the ASR increases from 10\% to almost 90\%. The same is true for small ($<$ 8\%) triggers in top-right (\autoref{fig:rate-vs-size-black-top-right} vs.~\autoref{fig:rate-vs-size-white-top-right}).
Additionally, for black and white triggers smaller than 8\% placed in the middle (\autoref{fig:rate-vs-size-black-middle} and~\autoref{fig:rate-vs-size-white-middle}), the ASR is low for all models except for AlexNet. In this case, the backdoor works only with a green-colored trigger.

\textbf{From all these observations, we conclude that the trigger color can play an important role in the backdoor's effectiveness, but it depends on many factors like the dataset or the model, making its optimization challenging for an attacker}.

\subsubsection{Effect of the Poisoning Rate}

Generally, with the increasing poisoning rate, all models' ASR increases. This is reasonable because, with more poisoned data, a backdoor attack can perform better. 
However, the attacker cannot increase $\epsilon$ indefinitely as the model's CAD is reduced when the poisoning rate grows. 

\begin{figure}
    \centering
    \includegraphics[width=0.85\linewidth]{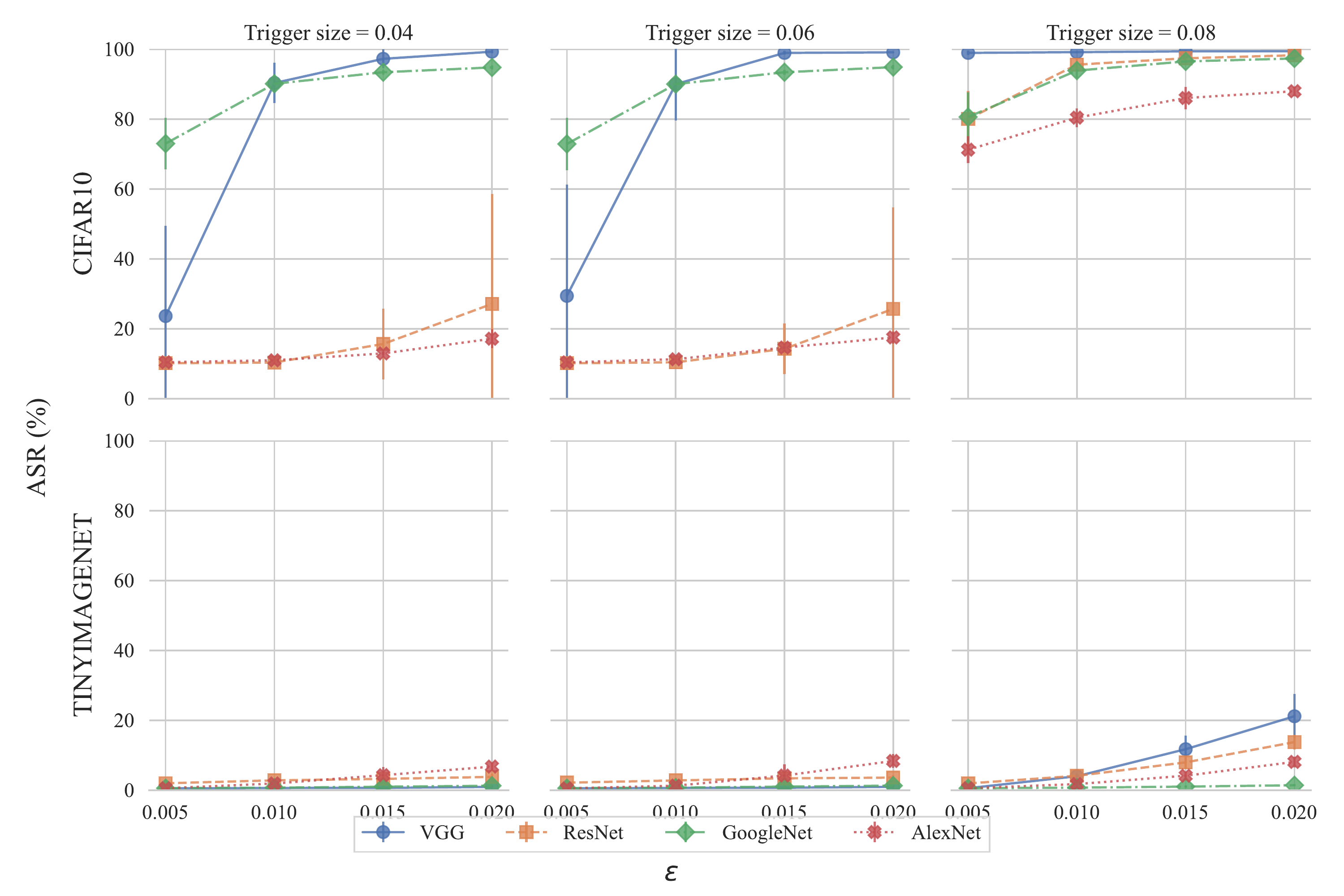}
    \caption{Rate vs. size, green color, trigger at bottom-right}
    \label{fig:rate-vs-size-green-bottom-right}
\end{figure}

\begin{table*}[htb]
\centering
\caption{Summary of the results. $\blackbox$ means that the ASR for at least one model is higher than 80\%, $\squarelrblack$ that the ASR is between 60\% and 80\%, and $\whitesquare$ that the ASR of every model is below 60\%.}
\label{tab:summary}
\resizebox{\linewidth}{!}{%
\begin{tabular}{@{}ccccccccccccccccc@{}}
\multirow{4}{*}{\backslashbox{Trigger position}{Trigger color}} 
    
    & \multicolumn{6}{c}{White} 
    & \multicolumn{6}{c}{Black}
    & \multicolumn{4}{c}{Green} \\ 
    \cmidrule(l){2-17} 
    
    & \multicolumn{12}{c}{Trigger size} \\
    & \multicolumn{3}{c}{0.04}   
    & \multicolumn{3}{c}{0.08}
    & \multicolumn{3}{c}{0.04}   
    & \multicolumn{3}{c}{0.08}
    & \multicolumn{2}{c}{0.04}
    & \multicolumn{2}{c}{0.08} \\
    \bottomrule
    
    &
    MNIST &
    CIFAR10 &
    \multicolumn{1}{c}{TinyImageNet} &
    MNIST &
    CIFAR10 &
    \multicolumn{1}{c}{TinyImageNet} &
    MNIST &
    CIFAR10 &
    \multicolumn{1}{c}{TinyImageNet} &
    MNIST &
    CIFAR10 &
    \multicolumn{1}{c}{TinyImageNet} &
    CIFAR10 &
    \multicolumn{1}{c}{TinyImageNet} &
    CIFAR10 &
    TinyImageNet \\
    
    Top-left
    &   $\blackbox$      & $\blackbox$       & $\whitesquare$
    &   $\blackbox$      & $\blackbox$       & $\whitesquare$ 
    &   $\whitesquare$   & $\blackbox$       & $\whitesquare$ 
    &   $\whitesquare$   & $\blackbox$       & $\whitesquare$ 
    
    &   $\blackbox$       & $\whitesquare$ 
    &   $\blackbox$       & $\whitesquare$  

    \\
    
    Middle
    &   $\whitesquare$   & $\squarelrblack$& $\blackbox$
    &   $\squarelrblack$ & $\blackbox$     & $\blackbox$ 
    &   $\whitesquare$   & $\whitesquare$  & $\blackbox$ 
    &   $\squarelrblack$ & $\blackbox$     & $\blackbox$
    
    &   $\blackbox$       & $\blackbox$ 
    &   $\blackbox$       & $\blackbox$    
    \\
    
    Bottom-right
    &   $\blackbox$      & $\blackbox$       & $\whitesquare$
    &   $\blackbox$      & $\blackbox$       & $\whitesquare$ 
    &   $\whitesquare$   & $\blackbox$       & $\whitesquare$ 
    &   $\whitesquare$   & $\blackbox$       & $\whitesquare$
    
    &   $\blackbox$      & $\whitesquare$ 
    &   $\blackbox$      & $\whitesquare$ 
    \\
    
    \bottomrule
\end{tabular}%
}
\end{table*}


\subsection{On the Interpretability of Backdoors}

Convolutional layers capture the spatial information, so the last convolutional layer is expected to achieve the best understanding of high-level semantics and detailed spatial information. Thus, the neurons of convolutional layers look for the class-specific semantics, e.g., capturing image parts relevant to the label ``dog''. Grad-CAM uses this information for obtaining an attention map given an image and a target class. Intuitively, one can imagine Grad-CAM attention maps as the critical parts for a model to classify an image for the target label. 
Grad-CAM has also been widely applied in the image backdoor domain to explain the behavior of the backdoor triggers~\cite{nguyen2020input, salem2022dynamic}. More precisely, we also leverage Grad-CAM to explain the importance of the trigger location and color. We use CIFAR10 as a test dataset to compare the attention of the backdoored models and the clean models for both clean and backdoored samples. We selected CIFAR10 because it is a perfect candidate since it contains large (upscaled) color images, which is also representative of TinyImageNet and richer in features than MNIST. 
We select the setting from a successful backdoor attack to ensure that the trigger is getting injected. We experimented with a black trigger of size 8\% of the input image placed in the top-left corner. We set the $\epsilon$ value to 0.02 and train the models for 20 epochs. Simultaneously, we train a clean version of the same model and compute the attention maps for both clean and poisoned models. These maps show the image's most influential part (in red) for the model's output. Depending on the model used, we observe different behaviors. It is important to note that we use clean and target labels, i.e., the ground truth label and the backdoor label, to help understand the label's effect on the model's prediction. Intuitively, we expect a well-trained clean model to resist image perturbations (to some extent) as input triggers. Therefore, we expect the clean model's attention maps to look similar. However, on a backdoor model trained with clean and backdoor data, we expect to obtain a similar attention map (as the clean model's) for the clean images. Nevertheless, backdoor images should bring the model's attention toward the trigger.

In GoogLeNet, the clean model (see~\autoref{fig:googlenet_clean} in~\autoref{sec:appendix}) focuses on the center and center-right locations for clean and target labels. This effect also remains visible in the backdoor model (see~\autoref{fig:googlenet_bk} in~\autoref{sec:appendix}), caused by the backdoor ``idea'' where the attention on clean images does not vary. However, the backdoor model's attention drifts toward the trigger under its presence. In the clean model, the trigger is unnoticed.

In ResNet and VGG, we observe a similar, yet more evident behavior as in GoogLeNet. The clean model (see~\autoref{fig:resnet_clean} and~\autoref{fig:vgg_clean} in~\autoref{sec:appendix}) robustly resists the trigger presence without modifying the attention map and maintaining the same as the clean input. The backdoor model also focuses on the exact locations of the images, as the clean model does. On poisoned inputs, the backdoor model easily recognizes the presence of the trigger, directing attention toward it, see~\autoref{fig:resnet_bk} and~\autoref{fig:vgg_bk} in~\autoref{sec:appendix}.

AlexNet's attention maps are biased by the poor performance on the backdoor task. The heatmaps could intuitively help  explain it. AlexNet's predictions are based on observing all the areas from the image rather than focusing on a specific area, as done by the abovementioned models. Still, the clean model is robust against perturbations on the input, i.e., the attention map does not vary much, see~\autoref{fig:alexnet_clean} in~\autoref{sec:appendix}. Similarly, the backdoor model has a slightly different attention map on clean images. However, the model does not focus on the trigger but on the whole input space on backdoor images, see~\autoref{fig:alexnet_bk} in~\autoref{sec:appendix}.

\subsection{Advanced Triggers}
\begin{figure}[htb]
    \begin{subfigure}[t]{0.23\textwidth}
        \centering
        \includegraphics[width=0.95\textwidth]{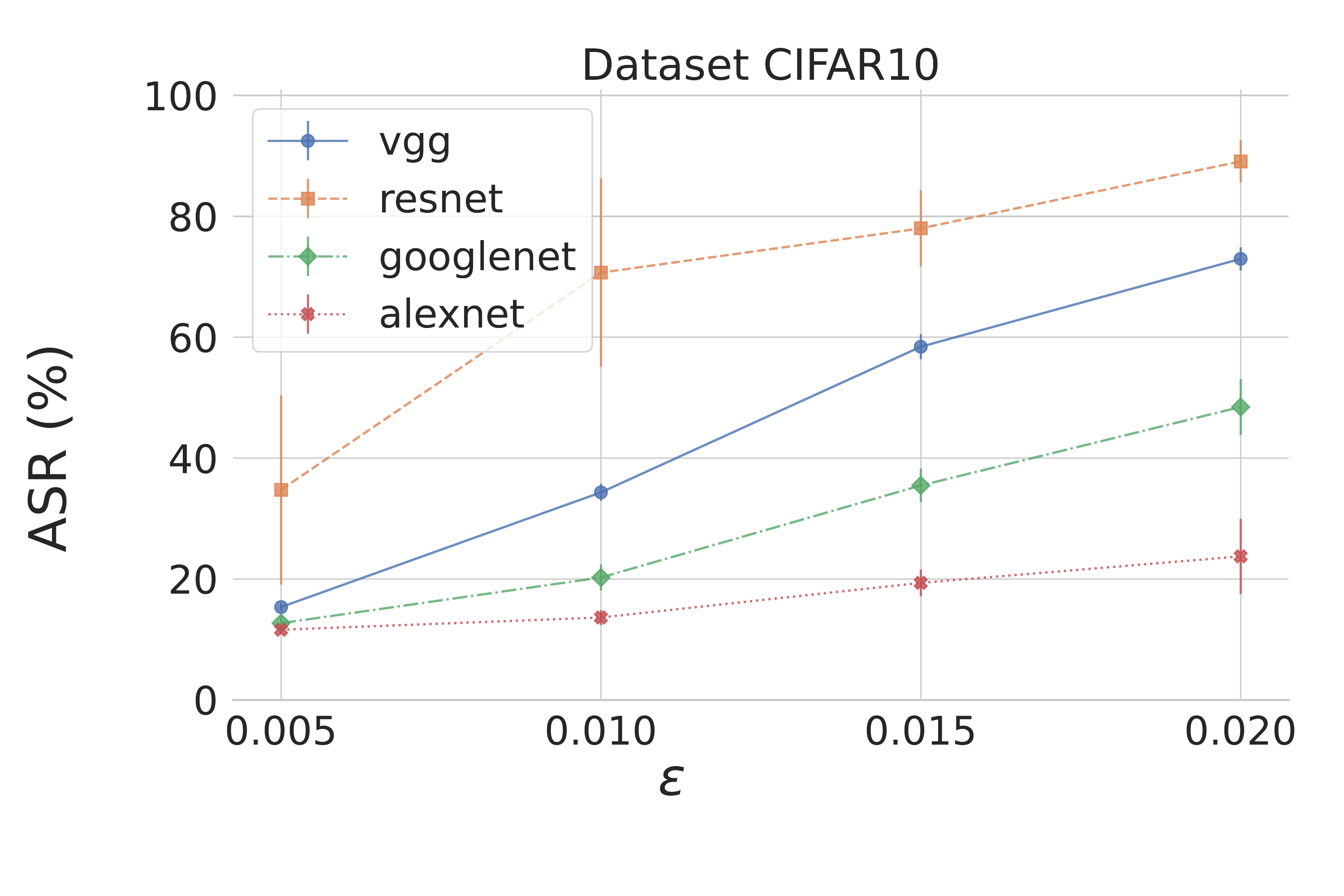}
        \caption{SSBA}
    \end{subfigure}
    \hfill
    \begin{subfigure}[t]{0.23\textwidth}
        \centering
        \includegraphics[width=0.95\textwidth]{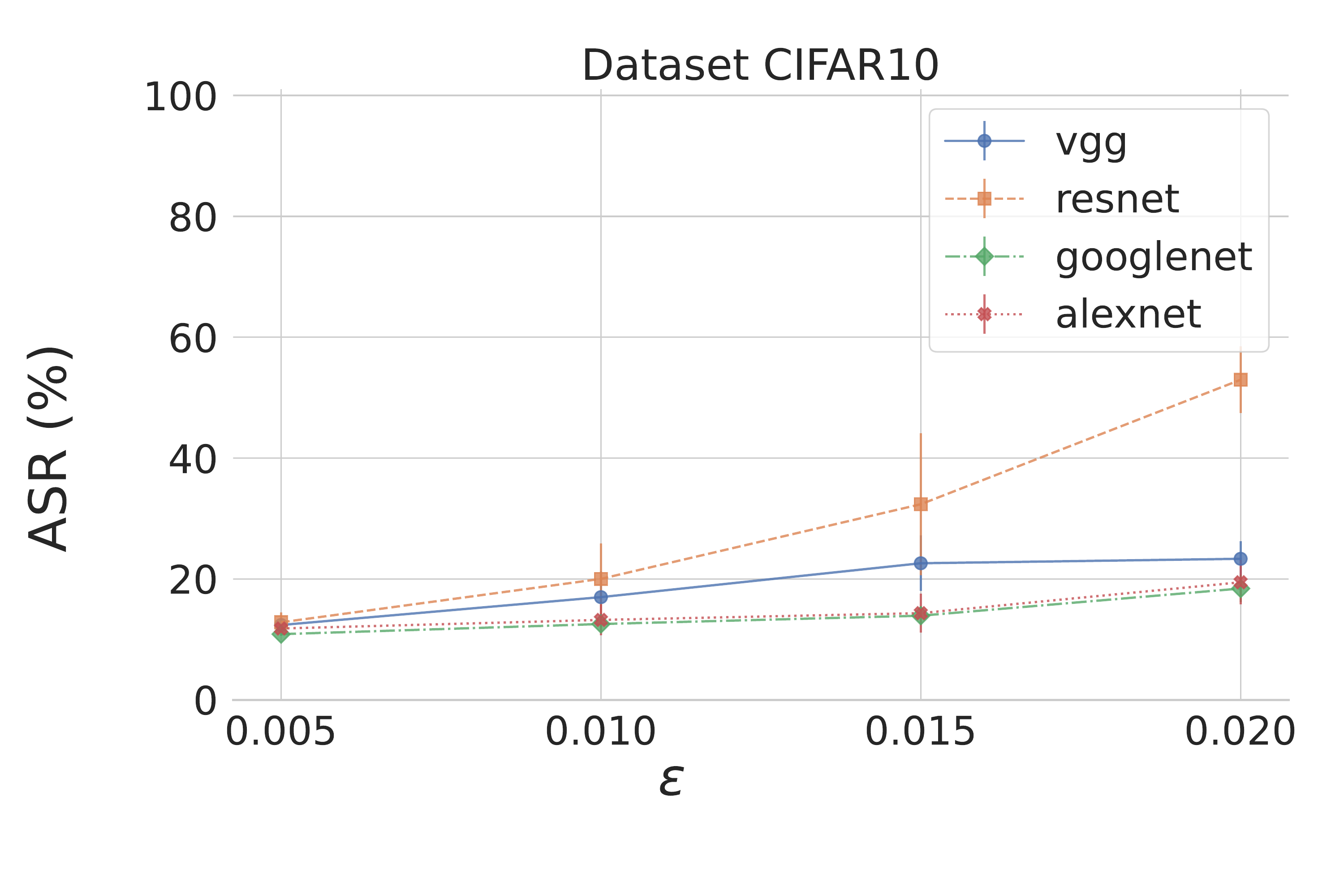}
        \caption{WaNet}
    \end{subfigure}
    \caption{ASR vs $\epsilon$ for SSBA and WaNet attacks.}
    \label{fig:asr_ssba_wanet}
\end{figure}

In this section, we investigate the efficacy of more subtle triggers compared to BadNets, with a specific focus on WaNet~\cite{wanet-imperceptible-warping-based-backdoor-attack} and SSBA~\cite{ssba}. After a backdoor was inserted with these two attacks there was no substantial performance drop on the original task for the poisoned models, similar to the BadNets attack.
The results from SSBA and WaNet are presented in \autoref{fig:asr_ssba_wanet}.
Notably, both SSBA and WaNet utilize triggers as large as the image itself, and we observe a clear correlation between the ASR and the poisoning rate for these attacks. However, unlike the BadNets attack, a higher poisoning rate is required to successfully inject the backdoor into the model using SSBA and WaNet. For instance, at the highest $\epsilon$ value tested, SSBA achieves a maximum ASR of 90\%, while WaNet achieves 59\%. In contrast, BadNets achieves 99\% ASR under the same experimental conditions.

Moreover, we observe that a weaker attacker with limited access to a smaller number of data samples would be unable to inject the backdoor using SSBA or WaNet. Specifically, with $\epsilon = 0.005$, which constitutes only 0.5\% of the available dataset, SSBA and WaNet achieve low ASR, obtaining only up to 50\% and 12\% ASR, respectively. This suggests that the viability of these ``advanced triggers'' may be limited in real-world scenarios where the attacker has limited access to data samples and transfer learning is used.

It is worth mentioning that the primary objective of these ``advanced trigger'' is to remain inconspicuous and imperceptible to human inspection, underscoring the stealthy nature of these attacks. Furthermore, the practical applicability of these attacks may be constrained in real-world scenarios where the attacker has limited access to data samples.



\color{black}

\subsection{Discussion}









We discuss several aspects of the backdoor attacks in image classification based on our experimental findings. First, we saw that the backdoor attack is easier when training from scratch. Thus, in future works, authors claiming that their trigger generation technique is stronger than the state-of-the-art should also run experiments in a transfer learning setup.
\begin{remark}
    The backdoor attack is easier when training from scratch. 
\end{remark}

Additionally, we should always use small poisoning rates as the clean accuracy drop increases when the poisoning rate is increased. Our experiments indicate that the drop is more severe for stronger models and larger datasets. Thus, there is no guarantee of a small clean accuracy drop in large datasets if we see no clean accuracy drop in small datasets.

\begin{remark}
    The clean accuracy drop increases as the poisoning rate increases. Additionally, we conjecture that the drop can be more severe for large datasets and strong models.
\end{remark}

We also saw that models with a large capacity and a large number of weights are more vulnerable to backdoor attacks. These models can overfit to small subsets of their datasets and learn complex patterns even from only a handful of training samples.

\begin{remark}
    Large models with big capacities are more vulnerable to backdoor attacks.
\end{remark}

From our experiments, we saw that no position, color, or combination of them results in the most effective backdoor across all settings. The best trigger color and position for every setup depends on the dataset, and the model used.

\begin{remark}
    No position or color results in the most effective backdoor universally. 
\end{remark}

%
%

Another observation from our experiments is that the ASR can vary for different trigger positions. Even though CNNs should not be affected by the feature (trigger) position, it seems that in some cases, they exploit the feature's absolute spatial location and learn the trigger easier. This was also shown in~\cite{on-translation-invariance-in-cnns-convolutional-layers-can-exploit-absolute-spatial-location} but not in the context of backdoor attacks. 
\begin{remark}
    The backdoor's performance varies for different trigger positions indicating that in some cases, the CNNs exploit the absolute spatial location of their features.
\end{remark}

As shown in~\autoref{fig:rate-vs-size-black-middle} for TinyImageNet, variations in the poisoning rate show improvement when the trigger size is 4\%. However, when the trigger is large, the $\epsilon$ does not affect much the backdoor performance. A similar effect is visible in~\autoref{fig:rate-vs-size-white-middle}, where with trigger size 0.04, variations in $\epsilon$ can drastically increase the backdoor performance. However, the poisoning rate is nearly irrelevant when the trigger size is large. 

\begin{remark}
    The trigger size has a more significant contribution to the ASR than the poisoning rate.
\end{remark}

When comparing the efficiency of patch triggers, i.e., BadNets, with ``advanced triggers'', we observe an overall lower ASR. Based on the poisoning rate, we observe that ``advanced triggers'' may not be realistically applicable in real-life scenarios since the attack requires access to a large amount of data, which may not be accessible. The attack's success also depends on the model, in which ASR varies drastically. However, they create more subtle perturbation in the images; thus, depending on the countermeasures applied by the model owner (see~\autoref{sec:defenses}, ``advanced triggers'' could be a viable alternative.

\begin{remark}
    Although ``advanced triggers'' could not be viable in realistic scenarios, they are more stealthy than patch triggers. 
\end{remark}

%% file: sections/_defenses.tex
\section{On the Defenses}
\label{sec:defenses}

In this section, we evaluate the attacks against state-of-the-art defenses. First, we briefly discuss the existing countermeasures. Then, we explain the chosen methods in more detail. Lastly, we evaluate the attacks against chosen countermeasures.

\subsection{Discussion}

Several defense mechanisms have been proposed in the literature for mitigating backdoor attacks. Neural Cleanse (NC)~\cite{wang2019neural} is the first to evaluate possible defense mechanisms by proposing a method that reverses engineers the trigger for each class, while outliers in the \emph{L1 norm} suggest the model has been compromised. However, the optimization process to reverse engineer the trigger is costly. It has to repeatedly be done for each label, which can be unfeasible for datasets with many classes. Recent research has shown that multi-triggers, large triggers, or input-specific labels can easily bypass NC.
Similarly, ABS~\cite{liu2019abs} has shown improved performance by improving NC. ABS stimulates neurons in the network and examines the outputs for deviations from the expected behavior. Authors suggest that a class can be represented as a subspace within a feature space, and a backdoor class will therefore create a distinct subspace in the feature space. ABS then relies on the fact that compromised neurons will produce larger outputs than no compromised neurons.

Following another approach, Liu et al.~\cite{liu2018fine} proposed combining neuron pruning and fine-tuning. Fine-pruning is a post-training defense that prunes the most ``active'' neurons and then fine-tunes the network for some epochs. The intuition is based on the fact that some neurons contain the main (clean) task information, others the backdoor task, and the rest a combination of both. Therefore, removing the correct group of neurons will reduce the backdoor effect. Sometimes, pruning is unnecessary, solely fine-tuning is enough to reduce the backdoor effect while maintaining high accuracy in the main task~\cite{abad2023sneaky}.

\subsection{Experimentation}

We evaluate the attacks against the most representative defense mechanisms, i.e., NC and fine-pruning. We evaluated the effect of these two defenses on the backdoor performance, in this section, we aim to research if the attacks with ``best'' hyperparameters are still robust against the chosen countermeasures. To be consistent between all the attacks, we select to perform our evaluation on the CIFAR10 dataset---as SSBA attack does not provide an assessment on MNIST dataset~\cite{invisible-backdoor-attacks-on-dnns-via-steganography-and-regularization}. 
For each attack, we selected the setup with the highest ASR. Specifically, for the BadNets attack, we selected a green patch placed in the center with size 8\% and a poisoning rate of 2\%. 
For the WaNet attack, we selected a poisoning rate of 0.02 for GoogLeNet, AlexNet, ResNet and a poisoning rate of 0.015 for VGG. 
For the SSBA attack, we selected the poisoning rate of 0.02 for all models.

NC evaluation against BadNets shows excellent performance on ResNet and AlexNet, successfully detecting the backdoored model and the target label. However, on GoogLeNet, the target label gets erroneously detected, although the model is correctly noticed as malicious. Lastly, on VGG, NC cannot detect the backdoor, i.e., the model is not flagged as malicious. 
Regarding WaNet, NC can always detect the backdoored model on ResNet and VGG, but it cannot accurately detect the target label. On GoogLeNet and AlexNet, NC is unable to detect the malicious model. 
For SSBA, NC successfully detects the backdoored models on AlexNet, ResNet, and VGG, although the target label is not precisely detected, i.e., more than one target label is identified. However, NC is unable to detect the backdoored model on GoogLeNet.

\begin{figure}[htb]
    \begin{subfigure}[t]{0.15\textwidth}
        \centering
        \includegraphics[width=0.95\textwidth]{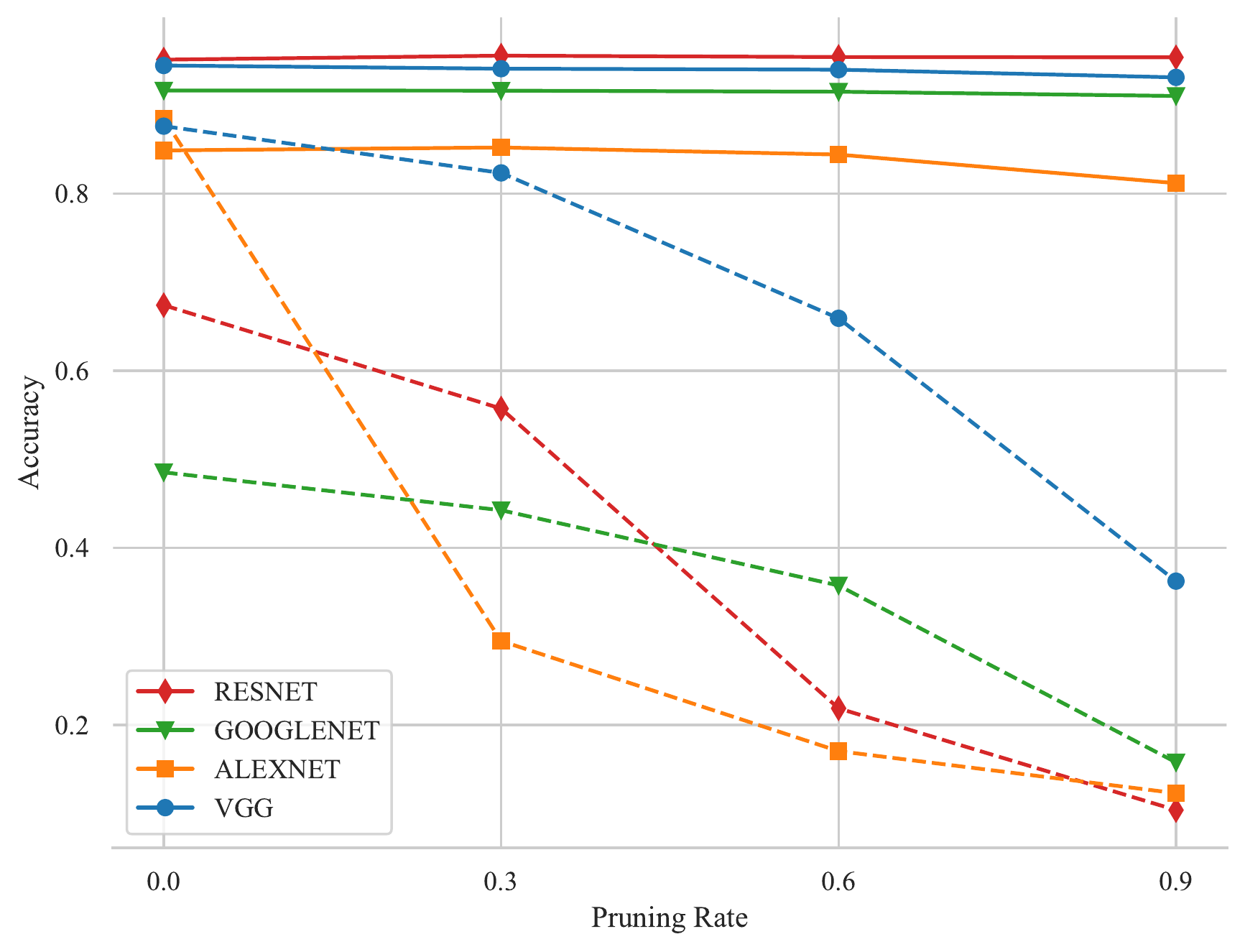}
        \caption{BadNets}
    \end{subfigure}
    \hfill
    \begin{subfigure}[t]{0.15\textwidth}
        \centering
        \includegraphics[width=0.95\textwidth]{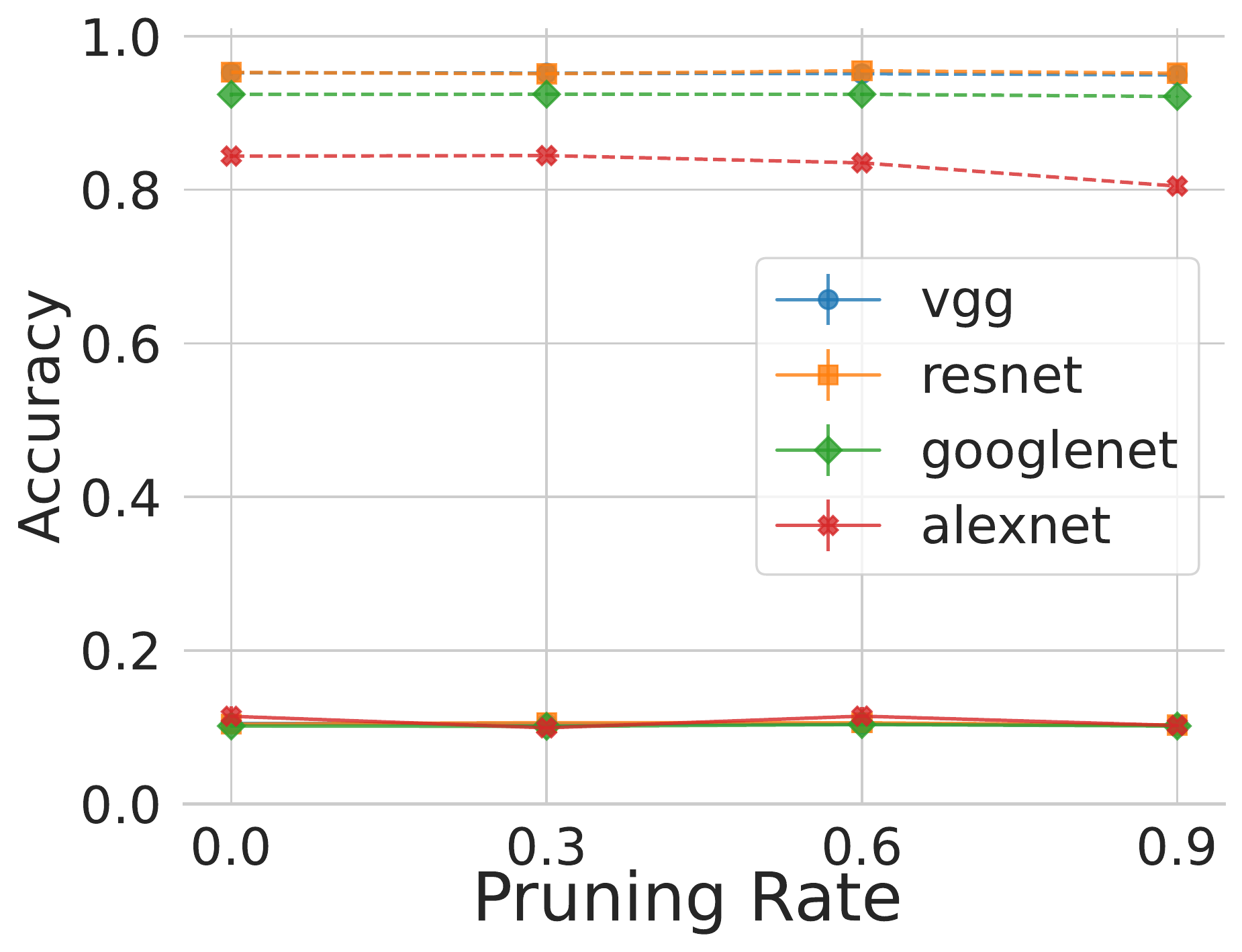}
        \caption{WaNet}
    \end{subfigure}
    \hfill
    \begin{subfigure}[t]{0.15\textwidth}
        \centering
        \includegraphics[width=0.95\textwidth]{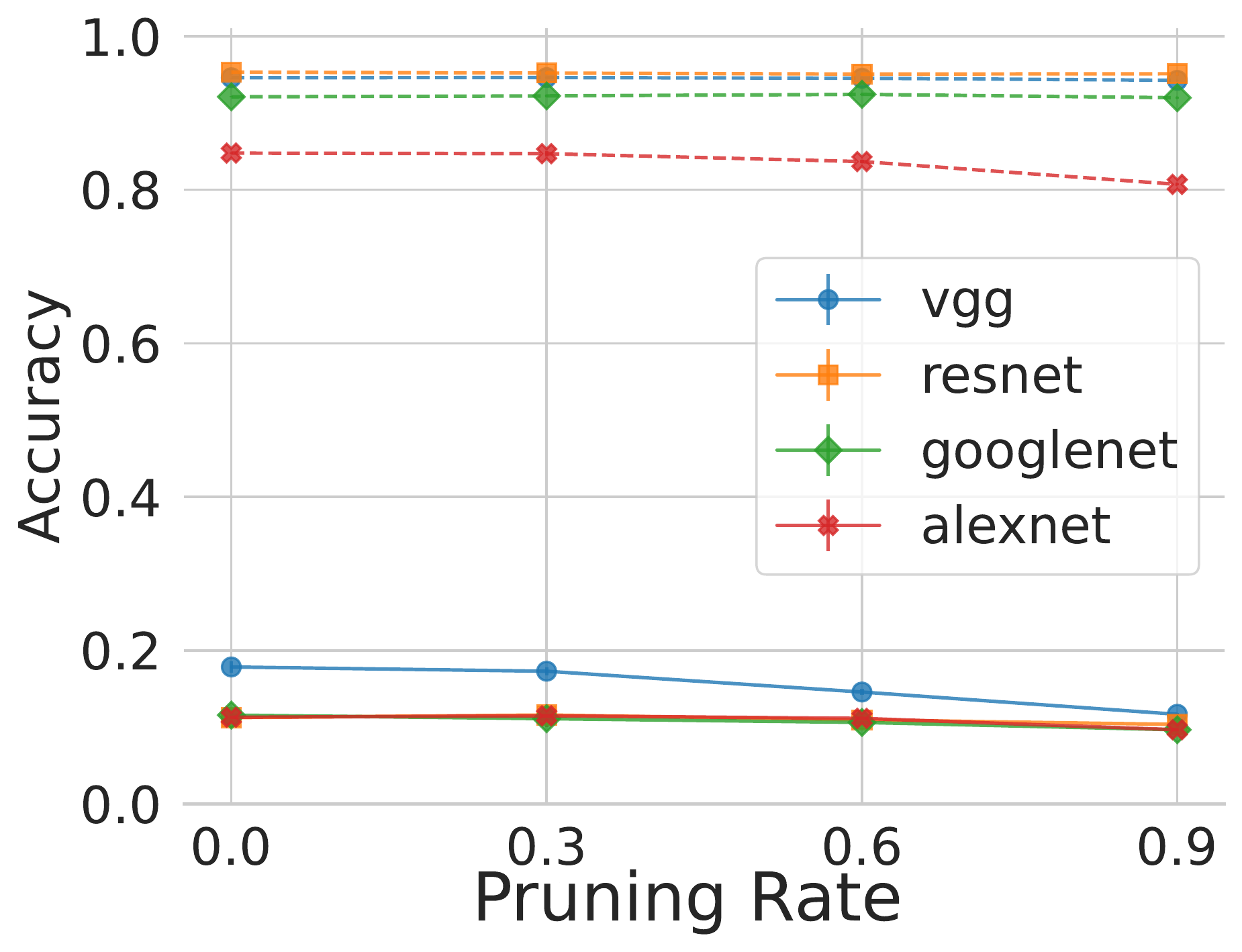}
        \caption{SSBA}
    \end{subfigure}
    \caption{Fine-Pruning against BadNets, WaNet, and SSBA (dashed and solid lines are the ASR and clean testing accuracy, respectively).}
    \label{fig:fine-pruning}
\end{figure}

We also evaluate the attacks against fine-pruning, see~\autoref{fig:fine-pruning}. We selected different pruning rates, i.e., the number of neurons to prune, and retrained the model for 10\% of the used initial epochs, which is a common practice. The BadNets attack is easily mitigated without a noticeable drop in clean accuracy. Moreover, when increasing the pruning rate to 90\%, we observe an almost complete elimination of the backdoor behavior. 
As for the SSBA and WaNet attacks, even with a pruning rate of 0\%, the ASR decreases dramatically for both attacks, which means that SSBA and WaNet are not even robust against fine-tuning. Moreover, as the pruning rate increases, the ASR remains low, i.e., around 10\%.



%% file: sections/5_related_work.tex
\section{Related Work}
\label{sec:related_work}

\begin{table*}
\centering
\caption{Comparison of the considered parameters in related works. ``Fixed'' means that the trigger position is not defined but fixed for all the experiments.}
\label{tab:comparison-related-work}
\resizebox{\linewidth}{!}{%
\begin{tabular}{@{}cccccccll@{}}
\toprule
     & Datasets & Models & Trigger color & Trigger shape & Trigger size & Trigger position & Trigger opacity & Poisoning rate \\ \midrule
     
Truong et al.~\cite{systematic-evaluation-of-backdoor-data-poisoning-attacks-on-image-classifiers}
&   \begin{tabular}[c]{@{}c@{}}Flowers~\cite{tfflowers}\\ CIFAR10\end{tabular}
&   \begin{tabular}[c]{@{}c@{}}ResNet50\\ NasNet~\cite{zoph2018learning}\\ NasNet Mobile~\cite{zoph2018learning}\end{tabular}
&   Black
&   \begin{tabular}[c]{@{}c@{}} Square\\ Overlay\end{tabular}
&   \begin{tabular}[c]{@{}c@{}} 22 pixels\end{tabular}
&   \begin{tabular}[c]{@{}c@{}} Top-left\\ Overlay\end{tabular}
&   Considered
&   \begin{tabular}[c]{@{}c@{}} 1\%\\ 2\%\\ 5\%\\ 10\%\\ 15\%\\\end{tabular} 
    \begin{tabular}[c]{@{}c@{}} 20\%\\ 25\%\\ 50\%\\ 75\%\\ 100\%\end{tabular}
\\
\\
Rehman et al.~\cite{rehman2019backdoor}
&   \begin{tabular}[c]{@{}c@{}}Belgian traffic signs~\cite{belgianradu2015van}\\ Chinese traffic signs~\cite{chinese2019}\\ French traffic signs~\cite{frenchpaparoditis2012stereopolis}\\ German traffic signs~\cite{germanstallkamp2012man}\end{tabular}
&   \begin{tabular}[c]{@{}c@{}}CNN\end{tabular}
&   \begin{tabular}[c]{@{}c@{}}White\\ Yellow\end{tabular}
&   \begin{tabular}[c]{@{}c@{}}Square\\ Star\end{tabular}
&   \begin{tabular}[c]{@{}c@{}} Not\\ considered\end{tabular}
&   \begin{tabular}[c]{@{}c@{}} Fixed\end{tabular}
&   \begin{tabular}[c]{@{}c@{}} Not\\ considered\end{tabular} 
&   \begin{tabular}[c]{@{}c@{}} 1\%\\ 3\%\\ 5\%\\ 10\%\\ 12.5\%\\ 15\%\end{tabular} 
\\
\\
\textbf{Ours}
&   \begin{tabular}[c]{@{}c@{}}MNIST\\ CIFAR10\\ TinyImageNet\end{tabular}
&   \begin{tabular}[c]{@{}c@{}}AlexNet\\ VGG\\ GoogLeNet\\ ResNet-152\end{tabular}
&   \begin{tabular}[c]{@{}c@{}}White\\ Black\\ Green\\ Dynamic\end{tabular}
&   \begin{tabular}[c]{@{}c@{}}Square\\ Whole Image\end{tabular}
&   \begin{tabular}[c]{@{}c@{}} 4\%\\ 6\%\\ 8\%\end{tabular}
&   \begin{tabular}[c]{@{}c@{}} Top-right\\ Top-left\\ Middle\\ Bottom-right\\ Bottom-left\end{tabular}
&   \begin{tabular}[c]{@{}c@{}}Not\\ considered\end{tabular}
&   \begin{tabular}[c]{@{}c@{}}0.5\%\\ 1\%\\ 1.5\%\\ 2\%\\\end{tabular}
\\ \bottomrule
\end{tabular}%
}
\end{table*}

\begin{table}[!htb]
\centering
\caption{Comparison of the effect of the parameters in related works. Where $\blackbox$ means completely considered, $\squarelrblack$ somehow considered, and $\whitesquare$ not considered.}
\label{tab:comparison-effect}
\resizebox{\linewidth}{!}{%
\begin{tabular}{@{}ccccccccc@{}}
\toprule
    & \begin{tabular}[c]{@{}c@{}}Effect of the\\ poisoning rate\end{tabular} 
    & \begin{tabular}[c]{@{}c@{}}Effect of the\\ trigger size\end{tabular} 
    & \begin{tabular}[c]{@{}c@{}}Effect of the\\ trigger color\end{tabular} 
    & \begin{tabular}[c]{@{}c@{}}Effect of the\\ opacity\end{tabular} 
    & \begin{tabular}[c]{@{}c@{}}Effect of the\\ position\end{tabular} 
    & \begin{tabular}[c]{@{}c@{}}Effect of the\\ trigger types\end{tabular}
    & \begin{tabular}[c]{@{}c@{}}Explainability\end{tabular} 
    & \begin{tabular}[c]{@{}c@{}}Countermeasures\end{tabular} 

    \\ 
    \midrule
     
Truong et al.~\cite{systematic-evaluation-of-backdoor-data-poisoning-attacks-on-image-classifiers}
&   $\squarelrblack$
&   $\whitesquare$
&   $\whitesquare$
&   $\squarelrblack$
&   $\whitesquare$
&   $\blackbox$
&   $\whitesquare$
&   $\squarelrblack$
\\
\\
Rehman et al.~\cite{rehman2019backdoor}
&   $\whitesquare$
&   $\whitesquare$
&   $\whitesquare$
&   $\whitesquare$
&   $\whitesquare$
&   $\whitesquare$
&   $\whitesquare$
&   $\whitesquare$
\\
\\
\textbf{Ours}
&   $\blackbox$
&   $\blackbox$
&   $\blackbox$
&   $\whitesquare$
&   $\blackbox$
&   $\whitesquare$
&   $\squarelrblack$
&   $\blackbox$
\\ \bottomrule
\end{tabular}%
}
\end{table}

Backdoor attacks have been widely investigated in different domains in recent years. BadNets~\cite{badnets-evaluating-backdoor-attacks-on-dnns} was the first paper to address backdoor attacks in computer vision for image classification. Since then, backdoors have also been applied to different domains such as audio~\cite{koffas2021can,backdoor-attack-against-speaker-verification,koffas2022going}, graph neural networks~\cite{zhang2021backdoor, xu2021explainability}, spiking neural networks~\cite{abad2022poster}, natural language processing~\cite{blind-backdoors-in-deep-learning-models,badnl,tminer}, or collaborative learning~\cite{bagdasaryan2020backdoor,abadsniper, abad2021sok}. Specific to the image domain, different approaches have arisen: multi-trigger~\cite{kwon2020multi}, dynamic~\cite{nguyen2020input, salem2022dynamic, gap}, or invisible backdoors~\cite{li2021invisible,invisible-backdoor-attacks-on-dnns-via-steganography-and-regularization}, to name a few. 

At the same time, the security of ML concern grew, and the research community began investigating defense mechanisms to palliate this threat~\cite{backdoor-learning-a-survey,backdoor-attacks-and-countermeasures-on-deep-learning-a-comprehensive-review}. Most of these works include ablation studies that show the effects that various parameters have on the backdoor's effectiveness. However, the values used are different each time, which makes it challenging to compare the performance of different attacks. 

Fixing a parameter while evaluating the rest could provide insightful information about a single parameter. However, in ablation studies, how parameters combine is not evaluated~\cite{systematic-evaluation-of-backdoor-data-poisoning-attacks-on-image-classifiers, badnets-evaluating-backdoor-attacks-on-dnns}, which is indeed what defines the backdoor performance. Therefore, to understand which parameter is the most influential in the backdoor performance, both individual and combined parameters evaluation has to be done.


In this work, we focus on computer vision for image classification, the most popular application in the literature, and systematically evaluate the effect of various factors on the backdoor. Moreover, we find the most influential parameters by comparing their impact on the backdoor's effectiveness. 

Not many systematic evaluations have been done that study the effect of different parameters individually and together to discover their impact on backdoor success. To the best of our knowledge,~\cite{systematic-evaluation-of-backdoor-data-poisoning-attacks-on-image-classifiers, rehman2019backdoor} are two works that made some notable evaluations in the image domain in a systematic manner. 

In~\cite{systematic-evaluation-of-backdoor-data-poisoning-attacks-on-image-classifiers}, a similar work, the authors kept the number of samples for each class equal to avoid any dataset biases. We followed a more straightforward method that replaces clean samples with their poisoned counterparts and their changed labels because it is more prevalent in the literature~\cite{badnets-evaluating-backdoor-attacks-on-dnns,targeted-backdoor-attacks-on-deep-learning-systems-using-data-poisoning,backdoor-embedding-in-cnns-via-invisible-perturbation,bypassing-backdoor-detection-algorithms-in-deep-learning,composite-backdoor-attack-for-dnn-by-mixing-benign-features}. Additionally, we used datasets with larger images and more classes to explore if the observed behavior can be generalized for different settings. Furthermore, based on our results, we extract model/dataset-specific observations leveraging more generalized findings.

In~\autoref{tab:comparison-related-work}, we compared the parameters of previous works considered for their investigations. We found that neither of the previous works has performed a thorough evaluation. Precisely, in~\cite{systematic-evaluation-of-backdoor-data-poisoning-attacks-on-image-classifiers} only considered two datasets with the same number of classes and three models. Although the backdoor attack with two different trigger shapes (a square and an overlay) has been considered, only a single trigger color has been used. Contrary to our work, they considered different trigger opacities.

Nevertheless, we find their chosen trigger size (only one setting) and their selection of poisoning rates unrealistic. Indeed, the poisoning rate should be maintained small, as the attacker cannot access a large part of the training set. In~\autoref{tab:comparison-effect}, we analyze what parameter effects have been considered in previous work. Truong et al.~\cite{systematic-evaluation-of-backdoor-data-poisoning-attacks-on-image-classifiers} compared the effect of the trigger types in detail by comparing the effects of square, sine, and variance triggers. However, the effect of the poisoning rate, trigger opacity, and regularization as a defense mechanism has not been wholly evaluated (they have only been tested for a specific setting). Lastly, evaluation of the trigger size, color, position, and backdoor explainability are missing.

The investigation performed by Rehman et al.~\cite{rehman2019backdoor} only considered traffic signs datasets, so results cannot be generalized to the broad image domain. Furthermore, the only consideration of a simple CNN is far for the real-world used DL models. Although their considered different trigger colors and shapes, different trigger positions are not evaluated, which could provide inaccurate results as the traffic signs datasets are usually centered. This could lead to a potential misunderstanding of the results. Contrary to previously analyzed work~\cite{systematic-evaluation-of-backdoor-data-poisoning-attacks-on-image-classifiers} and ours, the authors have not analyzed the effects of the chosen parameters, thus not providing any insight into what is more important for a backdoor trigger in the traffic sign domain, as see in~\autoref{tab:comparison-effect}.

Considering the previous evaluation and motivated by previous work, we investigated the encountered gaps in the evaluations. Also, based on the found experimental inconsistencies and to provide accurate and bias-less results, we performed 10,800 experiments containing all the models, datasets, and attack settings. Additionally, our further investigation on AlexNet was carefully performed over 1,800 trained models.

%% file: sections/6_conclusions.tex
\section{Conclusions and Future Work}
\label{sec:conclusions}

This study investigates the impact of backdoor parameters on image classification, intending to identify the most influential parameters for backdoor success. However, we observed that backdoor attacks exhibit heterogeneity, challenging direct comparisons. Therefore, we focused on a core subgroup of backdoor attacks based on the BadNets approach. We conducted an extensive literature review and devised a systematic experimental setup encompassing standard backdoor designs, allowing us to gain insights into which parameters significantly affect backdoor performance.
Our research fills a gap in the existing literature by providing model/dataset-specific findings, some of which may be generalizable. Specifically, our empirical findings shed light on i) injecting backdoors in realistic scenarios, such as transfer learning; ii) the reasoning behind the backdoor effect; and iii) efficient backdoor injection through parameter tuning.

Two key findings emerged from our study. First, we found that trigger size has a more significant impact on backdoor success than the poisoning rate. This has important implications for designing countermeasures against backdoor attacks, as larger trigger sizes are more relevant, contrary to previous work that focused on small triggers only~\cite{aytekin2022neural}. Second, we found that training a model from scratch facilitates more straightforward backdoor injection than transfer learning. This finding has implications for future attack and defense designs, where fine-tuning must be considered to offer a realistic perspective on the proposal.

This paper aims to contribute to the research community by providing a reference framework for systematically comparing backdoor attack baselines, enabling comparable and reproducible results. However, it is important to acknowledge some limitations of our study. For instance, considering other trigger parameters, such as shape or opacity, could yield more robust findings. Additionally, our focus on patch-based triggers may limit the generalizability of our findings to more complex attacks involving dynamic or blending backdoors. Furthermore, we did not evaluate the impact of defense mechanisms on the choice of backdoor parameters, which could yield interesting findings on the best parameters for defense evasion. Our research specifically focuses on identifying the best parameters for backdoor injection.

This study has raised several questions that require further investigation. For example, more research on backdoor explainability or interpretability would enhance the accuracy of our findings. Furthermore, additional studies are needed to establish a solid foundation for comparing other types of backdoor attacks, validating their performance, and pushing the research community towards more robust and comprehensible backdoor attacks. Specifically, future work could explore the following:

\begin{compactenum} 
    \item Different trigger shapes and types, such as dynamic or blending triggers. 
    \item Consideration of defense mechanisms and the stealthiness of backdoor triggers. 
    \item As our findings suggest, the optimizer may play a significant role in the performance of backdoor attacks. Therefore, further investigation could provide valuable insights for developing more robust models. 
\end{compactenum}

%% file: sections/7_appendix.tex
\appendix

\section{Additional Experiments and Results}
\label{sec:appendix}

\subsection{On the Effect of Freezing AlexNet Layers}

\begin{figure}
    \centering
    \includegraphics[width=\linewidth]
    {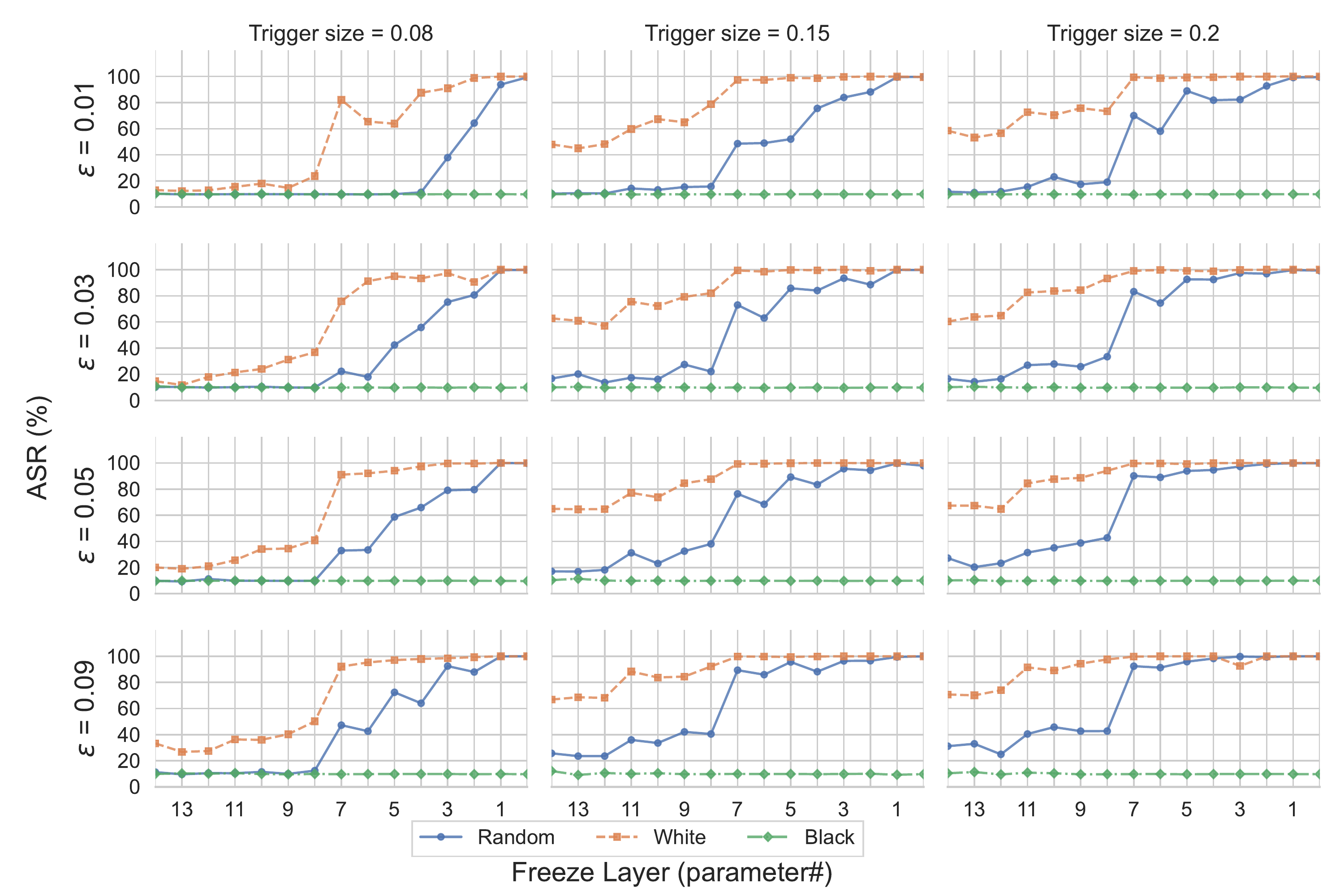}
    \caption{AlexNet on MNIST: FreezeLayer effect vs. size and rate, trigger at bottom-right}
    \label{fig:alexnet_mnist_freezelayer_sizerate_buttomright}
\end{figure}

As we discussed in~\Cref{sec:effect_of_model_architecture} we unfroze AlexNet parameters layer by layer (from 14 to 0) to see from which layer it starts to react positively on the injected backdoor. In~\autoref{fig:alexnet_mnist_freezelayer_sizerate_buttomright} we show the results for MNIST and we see that after the 7th parameter, the ASR is increased significantly.

\subsection{Features Maps}

\begin{figure}
\centering

\begin{subfigure}[b]{0.5\textwidth}
\centering
\includegraphics[width=.45\textwidth]{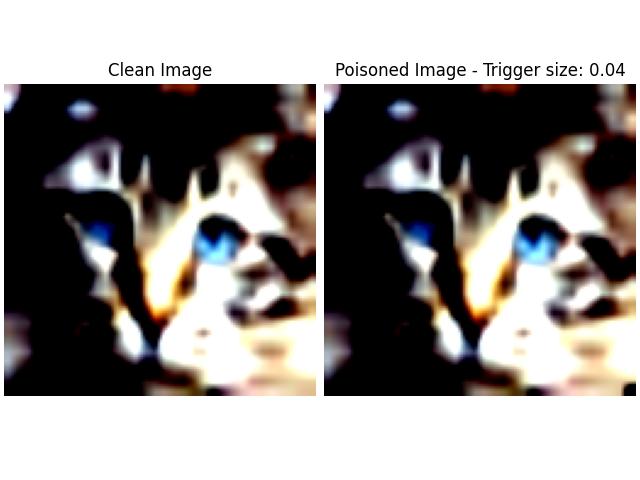}\quad
\includegraphics[width=.45\textwidth]{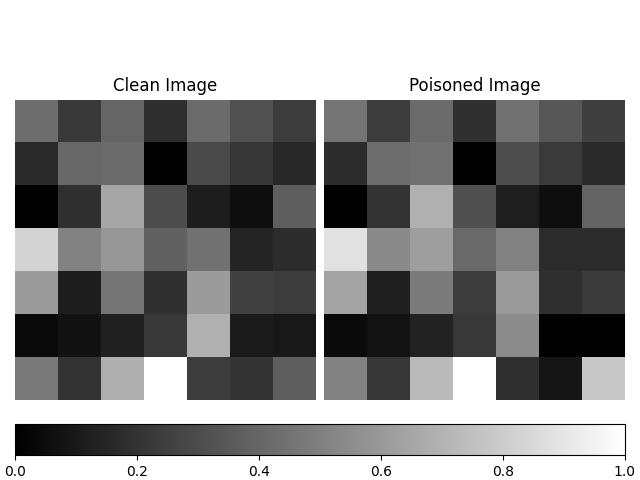}

\caption{AlexNet's features module (last Conv layer) output for a sample CIFAR10 image and its poisoned equivalent. Left: images $\rightarrow$ Right: feature maps. Trigger\_size=$0.04$, $\epsilon=0.01$, color = black, bottom-right (target label: airplane $\rightarrow$ model prediction: cat)}
\label{fig:cifar10_clanddif_size0.04}
\end{subfigure}\quad
\begin{subfigure}[b]{0.5\textwidth}
\centering
\includegraphics[width=.45\textwidth]{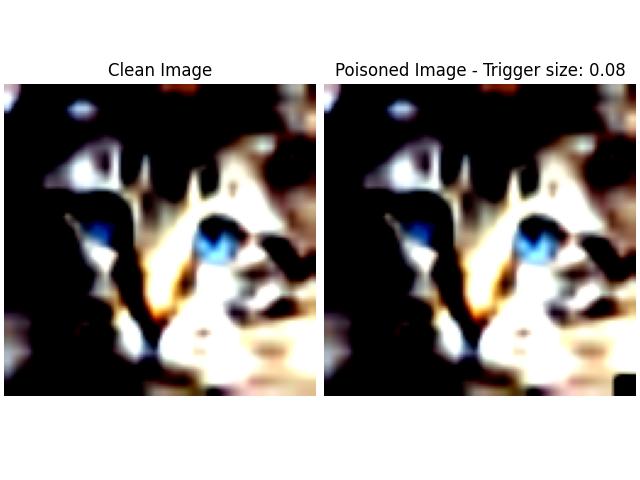}\quad
\includegraphics[width=.45\textwidth]{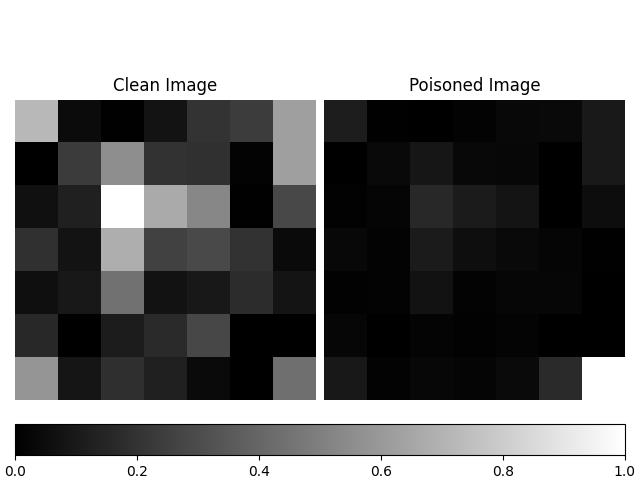}

\caption{AlexNet's features module (last Conv layer) output for a sample CIFAR10 image and its poisoned equivalent. Left: images $\rightarrow$ Right: feature maps. Trigger\_size=$0.08$, $\epsilon=0.01$,color = black, bottom-right (target label: airplane $\rightarrow$ model prediction: airplane)}
\label{fig:cifar10_clanddif_size0.08}
\end{subfigure}

\caption{AlexNet feature map for the same sample input and its poisoned equivalent. Consider the dissimilarity between the activations for two different sizes. The trigger size 0.08 can reach a high ASR, while the one with size 0.04 fails to get an ASR of more than 20\%}
\label{fig:AlexNet_fm_cifar10}

\end{figure}

Visualization of the feature map in the last convolutional layer for AlexNet can be seen in~\autoref{fig:cifar10_clanddif_size0.04} and~\autoref{fig:cifar10_clanddif_size0.08}. 

\subsection{Attention Maps}

\begin{figure}[!htb]
    \begin{subfigure}[t]{0.115\textwidth}
        \centering
        \includegraphics[width=0.95\textwidth]{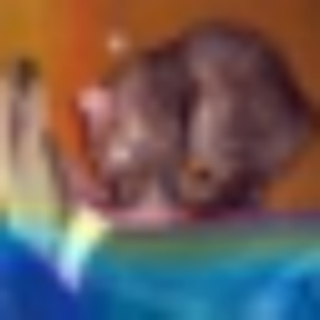}
        \caption{Clean input on the clean label}
    \end{subfigure}
    \hfill
    \begin{subfigure}[t]{0.115\textwidth}
        \centering
        \includegraphics[width=0.95\textwidth]{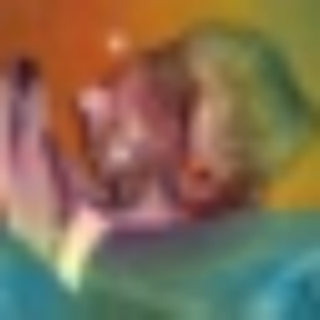}
        \caption{Clean input on the target label}
    \end{subfigure}
    \hfill
    \begin{subfigure}[t]{0.115\textwidth}
        \centering
        \includegraphics[width=0.95\textwidth]{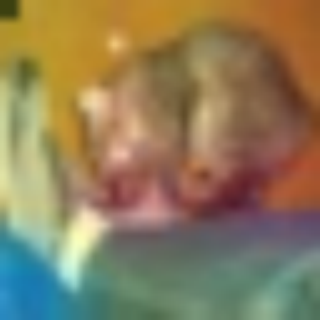}
        \caption{Backdoor input on the clean label}
    \end{subfigure}
    \hfill
    \begin{subfigure}[t]{0.115\textwidth}
        \centering
        \includegraphics[width=0.95\textwidth]{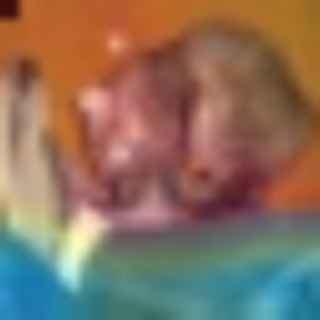}
        \caption{Backdoor input on the target label}
    \end{subfigure}
    \caption{GoogLeNet trained with clean data.}
    \label{fig:googlenet_clean}
\end{figure}

\begin{figure}[!htb]
    \begin{subfigure}[t]{0.115\textwidth}
        \centering
        \includegraphics[width=0.95\textwidth]{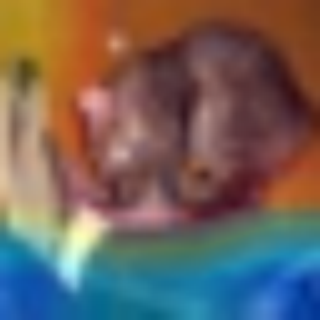}
        \caption{Clean input on the clean label}
    \end{subfigure}
    \begin{subfigure}[t]{0.115\textwidth}
        \centering
        \includegraphics[width=0.95\textwidth]{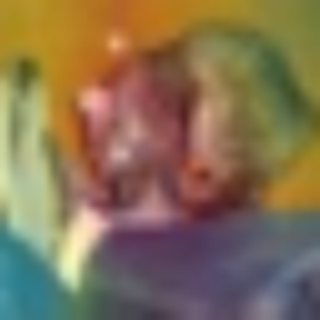}
        \caption{Clean input on the target label}
    \end{subfigure}
    \begin{subfigure}[t]{0.115\textwidth}
        \centering
        \includegraphics[width=0.95\textwidth]{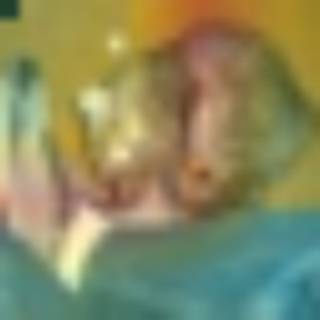}
        \caption{Backdoor input on the clean label}
    \end{subfigure}
    \begin{subfigure}[t]{0.115\textwidth}
        \centering
        \includegraphics[width=0.95\textwidth]{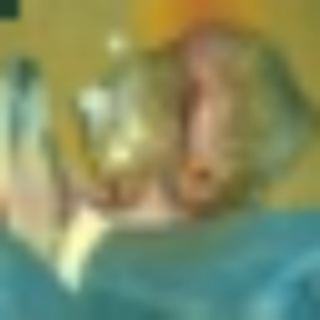}
        \caption{Backdoor input on the target label}
    \end{subfigure}
    \caption{GoogLeNet trained with poisoned data.}
    \label{fig:googlenet_bk}
\end{figure}

\begin{figure}[!htb]
    \begin{subfigure}[t]{0.115\textwidth}
        \centering
        \includegraphics[width=0.95\textwidth]{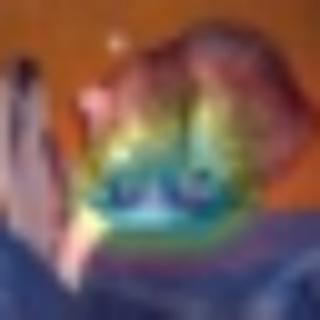}
        \caption{Clean input on the clean label}
    \end{subfigure}
    \begin{subfigure}[t]{0.115\textwidth}
        \centering
        \includegraphics[width=0.95\textwidth]{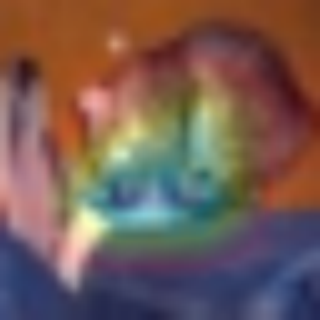}
        \caption{Clean input on the target label}
    \end{subfigure}
    \begin{subfigure}[t]{0.115\textwidth}
        \centering
        \includegraphics[width=0.95\textwidth]{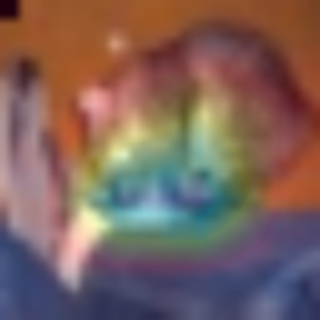}
        \caption{Backdoor input on the clean label}
    \end{subfigure}
    \begin{subfigure}[t]{0.115\textwidth}
        \centering
        \includegraphics[width=0.95\textwidth]{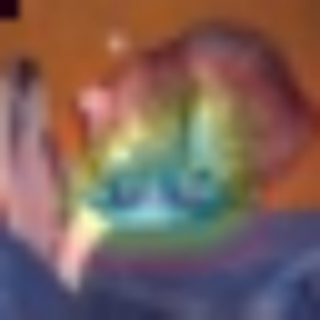}
        \caption{Backdoor input on the target label}
    \end{subfigure}
    \caption{ResNet trained with clean data.}
    \label{fig:resnet_clean}
\end{figure}

\begin{figure}[htb]
    \begin{subfigure}[t]{0.115\textwidth}
        \centering
        \includegraphics[width=0.95\textwidth]{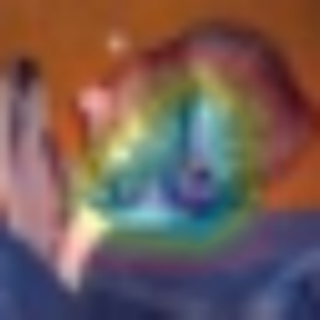}
        \caption{Clean input on the clean label}
    \end{subfigure}
    \begin{subfigure}[t]{0.115\textwidth}
        \centering
        \includegraphics[width=0.95\textwidth]{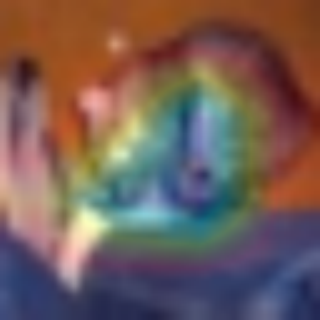}
        \caption{Clean input on the target label}
    \end{subfigure}
    \begin{subfigure}[t]{0.115\textwidth}
        \centering
        \includegraphics[width=0.95\textwidth]{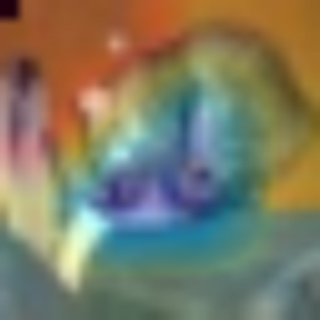}
        \caption{Backdoor input on the clean label}
    \end{subfigure}
    \begin{subfigure}[t]{0.115\textwidth}
        \centering
        \includegraphics[width=0.95\textwidth]{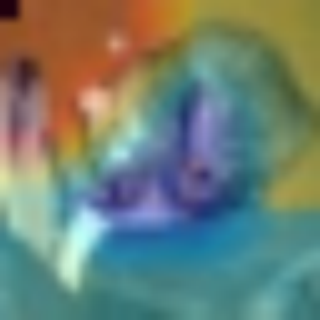}
        \caption{Backdoor input on the target label}
    \end{subfigure}
    \caption{ResNet trained with poisoned data.}
    \label{fig:resnet_bk}
\end{figure}

\begin{figure}[htb]
    \begin{subfigure}[t]{0.115\textwidth}
        \centering
        \includegraphics[width=0.95\textwidth]{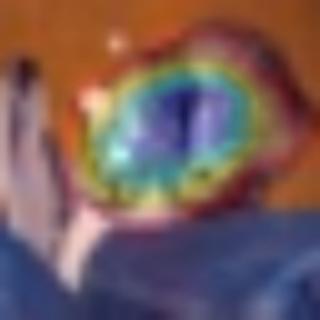}
        \caption{Clean input on the clean label}
    \end{subfigure}
    \begin{subfigure}[t]{0.115\textwidth}
        \centering
        \includegraphics[width=0.95\textwidth]{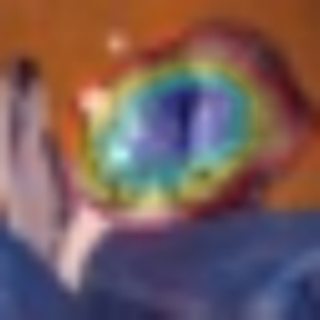}
        \caption{Clean input on the target label}
    \end{subfigure}
    \begin{subfigure}[t]{0.115\textwidth}
        \centering
        \includegraphics[width=0.95\textwidth]{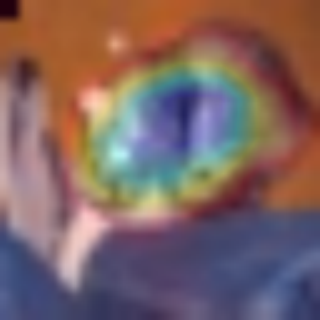}
        \caption{Backdoor input on the clean label}
    \end{subfigure}
    \begin{subfigure}[t]{0.115\textwidth}
        \centering
        \includegraphics[width=0.95\textwidth]{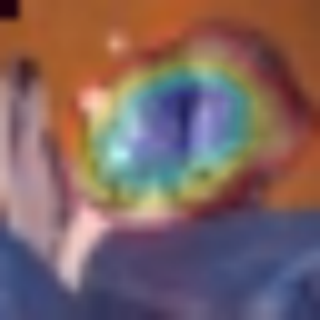}
        \caption{Backdoor input on the target label}
    \end{subfigure}
    \caption{VGG trained with clean data.}
    \label{fig:vgg_clean}
\end{figure}

\begin{figure}[htb]
    \begin{subfigure}[t]{0.115\textwidth}
        \centering
        \includegraphics[width=0.95\textwidth]{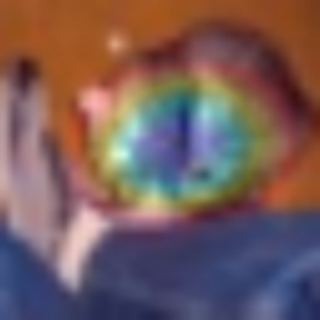}
        \caption{Clean input on the clean label}
    \end{subfigure}
    \begin{subfigure}[t]{0.115\textwidth}
        \centering
        \includegraphics[width=0.95\textwidth]{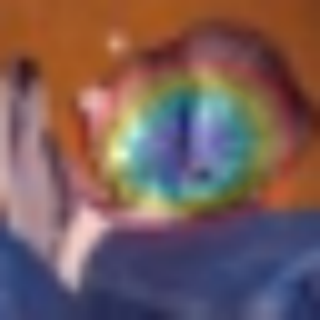}
        \caption{Clean input on the target label}
    \end{subfigure}
    \begin{subfigure}[t]{0.115\textwidth}
        \centering
        \includegraphics[width=0.95\textwidth]{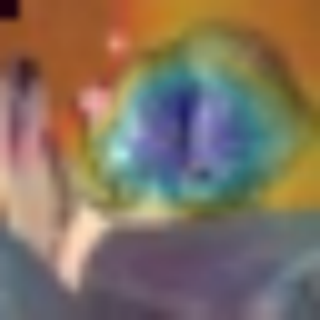}
        \caption{Backdoor input on the clean label}
    \end{subfigure}
    \begin{subfigure}[t]{0.115\textwidth}
        \centering
        \includegraphics[width=0.95\textwidth]{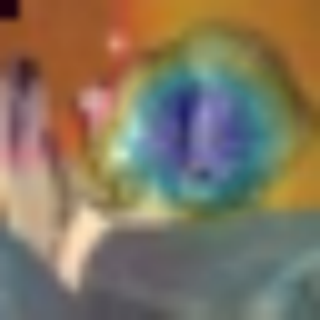}
        \caption{Backdoor input on the target label}
    \end{subfigure}
    \caption{VGG trained with poisoned data.}
    \label{fig:vgg_bk}
\end{figure}

\begin{figure}[htb]
    \begin{subfigure}[t]{0.115\textwidth}
        \centering
        \includegraphics[width=0.95\textwidth]{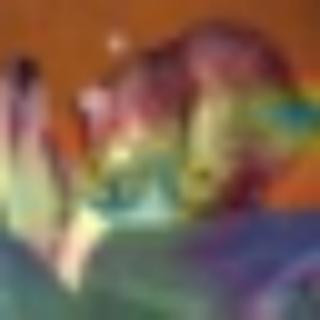}
        \caption{Clean input on the clean label}
    \end{subfigure}
    \begin{subfigure}[t]{0.115\textwidth}
        \centering
        \includegraphics[width=0.95\textwidth]{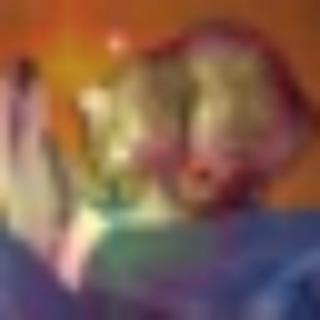}
        \caption{Clean input on the target label}
    \end{subfigure}
    \begin{subfigure}[t]{0.115\textwidth}
        \centering
        \includegraphics[width=0.95\textwidth]{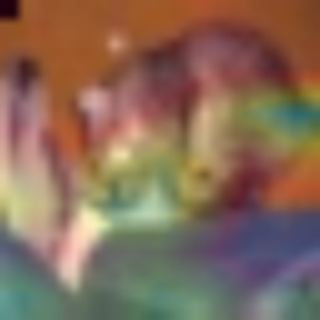}
        \caption{Backdoor input on the clean label}
    \end{subfigure}
    \begin{subfigure}[t]{0.115\textwidth}
        \centering
        \includegraphics[width=0.95\textwidth]{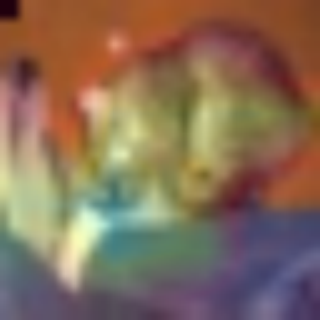}
        \caption{Backdoor input on the target label}
    \end{subfigure}
    \caption{AlexNet trained with clean data.}
    \label{fig:alexnet_clean}
\end{figure}

\begin{figure}[t]
    \begin{subfigure}[t]{0.115\textwidth}
        \centering
        \includegraphics[width=0.95\textwidth]{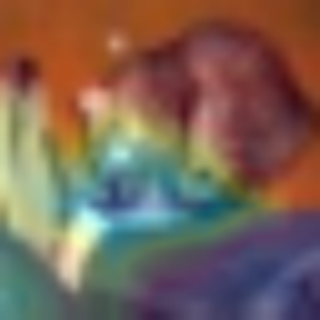}
        \caption{Clean input on the clean label}
    \end{subfigure}
    \begin{subfigure}[t]{0.115\textwidth}
        \centering
        \includegraphics[width=0.95\textwidth]{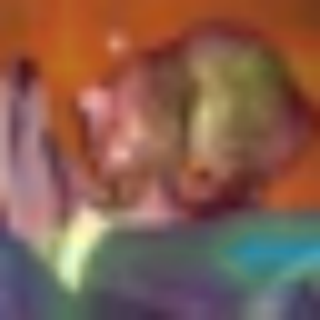}
        \caption{Clean input on the target label}
    \end{subfigure}
    \begin{subfigure}[t]{0.115\textwidth}
        \centering
        \includegraphics[width=0.95\textwidth]{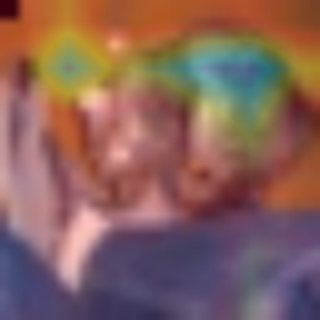}
        \caption{Backdoor input on the clean label}
    \end{subfigure}
    \begin{subfigure}[t]{0.115\textwidth}
        \centering
        \includegraphics[width=0.95\textwidth]{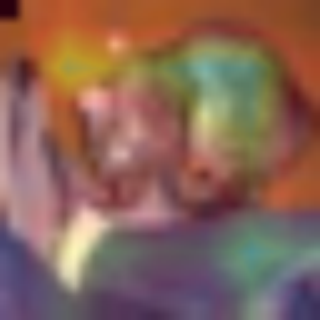}
        \caption{Backdoor input on the target label}
    \end{subfigure}
    \caption{AlexNet trained with poisoned data.}
    \label{fig:alexnet_bk}
\end{figure}

In this section, we show the generated attention maps for a backdoor model and the clean model.
We experimented with a black trigger of size 8\% of the input image placed in the top-left corner. We set the $\epsilon$ value to 0.02 and train the models for 20 epochs. The selected settings ensure a successful backdoor, i.e., the trigger is getting injected. In~\autoref{fig:googlenet_clean} and~\autoref{fig:googlenet_bk}, the clean and the backdoor attention maps for Googlenet are shown. In~\autoref{fig:resnet_clean} and~\autoref{fig:resnet_bk}, the clean and backdoor version of ResNet. In~\autoref{fig:vgg_clean} and~\autoref{fig:vgg_bk}, the backdoor version of VGG. And lastly, in~\autoref{fig:alexnet_clean} and~\autoref{fig:alexnet_bk}, the clean and backdoor version of AlexNet.

